\UseRawInputEncoding  %%for arxiv compile

\documentclass[review]{elsarticle}

\usepackage{lineno,hyperref}
\usepackage{framed,multirow}
\usepackage{amssymb}
\usepackage{latexsym}

\usepackage{url}
\usepackage{xcolor}

\usepackage{hyperref}
\usepackage{graphicx}
\usepackage{amsmath}
\usepackage{amssymb}
\usepackage{verbatim}
\usepackage{booktabs}
\usepackage{multirow}
\usepackage{bm}
\usepackage{array}
\usepackage{graphicx}
\usepackage{epstopdf}
\usepackage{subfigure}
\usepackage{amssymb}
\usepackage{amsmath}
\usepackage{makecell}
\usepackage{tabularx} % added
\usepackage{booktabs}   %three lines for table, added
\usepackage{mathrsfs} %added
\usepackage{multirow} %added
\usepackage{amsfonts} %added
\usepackage{wrapfig}  %added
\usepackage{cite}
\usepackage{footnote}
\usepackage{longtable}

\def\x{{\bf x}}

\def\0{{\bf 0}}
\def\1{{\bf 1}}

\def\etal{{\em et al.}}
\def\eg{{\em e.g.}}
\def\ie{{\em i.e.}}

\def\etal{{\em et al.\/}\,}

\graphicspath{{Figure/}}

\usepackage{geometry}
\begin{document}
	
	\begin{frontmatter}
		\title{Federated Learning for Medical Image Analysis: A Survey}

\author[mymainaddress]{Hao Guan}%\ead{guanhao@email.unc.edu}
\author[mymainaddress]{Pew-Thian Yap}
\author[mysecondaryaddress]{Andrea Bozoki}
\author[mymainaddress]{Mingxia~Liu\corref{mycorrespondingauthor}}%\ead{mxliu@med.unc.edu}
%\cortext[mycorrespondingauthor]{Corresponding author}\ead{dgshen@med.unc.edu}

\cortext[mycorrespondingauthor]{Corresponding author: mingxia\_liu@med.unc.edu}

\address[mymainaddress]{Department of Radiology and Biomedical Research Imaging Center, University of North Carolina at Chapel Hill, Chapel Hill, NC 27599, USA}

\address[mysecondaryaddress]{Department of Neurology, University of North Carolina at Chapel Hill, Chapel Hill, NC 27599, USA}

\begin{abstract}
%%%
Machine learning in medical imaging often faces a fundamental dilemma, namely, the small sample size problem. 
Many recent studies suggest using multi-domain data pooled from different acquisition sites/centers to improve statistical power. 
However, medical images from different sites cannot be easily shared to build large datasets for model training due to privacy protection reasons. 
As a promising solution, federated learning, which enables collaborative training of machine learning models based on data from different sites without cross-site data sharing, 
has attracted considerable attention recently. 
In this paper, we conduct a comprehensive survey of the recent development of federated learning methods in medical image analysis. 
We have systematically gathered research papers on federated learning and its applications in medical image analysis published between 2017 and 2023. Our search and compilation were conducted using databases from IEEE Xplore, ACM Digital Library, Science Direct, Springer Link, Web of Science, Google Scholar, and PubMed.  %
In this survey, we first introduce the background knowledge of federated learning for dealing with privacy protection and collaborative learning issues in medical imaging. We then present a comprehensive review of recent advances in federated learning methods for medical image analysis. Specifically, existing methods are categorized based on three critical aspects of a federated learning system, including client end, server end, and communication techniques. 
In each category, we summarize the existing federated learning methods according to specific research problems in medical image analysis and also provide insights into the motivations of different approaches. 
In addition, we provide a review of existing benchmark medical imaging datasets and software platforms for current federated learning research. 
We also conduct an experimental study to empirically evaluate typical federated learning methods for medical image analysis.
This survey can help to better understand the current research status, challenges, and potential research opportunities in this promising research field. %%%%
\end{abstract}

\begin{keyword}
%% MSC codes here, in the form: \MSC code \sep code
%% or \MSC[2008] code \sep code (2000 is the default)
%\MSC 41A05\sep 41A10\sep 65D05\sep 65D17
%% Keywords
Federated learning\sep machine learning\sep medical image analysis\sep data privacy
\end{keyword}

\end{frontmatter}

%\linenumbers
%%%%%%%%%%%%%%%%%%%%%%%%%%%%%%%%%%%%%%%%%%%%%%%%%%%%%%%%%%%%%%%%%%%%%%%%%%%%%%

%% main text
\section{Introduction} \label{Introduction}
Medical image analysis has been greatly pushed forward by computer vision and machine learning~\citep{barragan2021artificial,cheplygina2019not,guan2022domain,litjens2017survey}. 
The remarkable success of modern machine learning methods, \eg, deep learning~\citep{lecun2015deep}, can be attributed to the building and release of grand-scale natural image databases, such as ImageNet~\citep{Imagenet} and Microsoft Common Objects in Context (MS COCO)~\citep{COCO}. 
Unlike natural image analysis, the field of medical image analysis still faces the fundamental challenge of the ``small-sample-size" problem~\citep{raudys1991small,vabalas2019machine}.

\if false
In the field of medical imaging, however, a major challenge is the scarcity of labeled data because labeling medical images typically requires labor-intensive participation of doctors, radiologists, and other experts which is generally expensive, time-consuming, and tedious. 
\fi 
Based on small sample data, it is difficult for us to estimate real data distributions, greatly hindering the building of robust and reliable machine learning models for medical image analysis. 
An intuitive and direct solution to this small sample size problem is to pool images from multiple sites together and build larger datasets to train high-quality machine learning models. 
However, sharing medical imaging data between different sites/centers is intractable due to strict privacy protection policies such as Health Insurance Portability and Accountability Act (HIPAA)~\citep{HIPAA} and General Data Protection Regulation (GDPR)~\citep{GDPR}. 
For example, the United States HIPAA has rigidly restricted the exchange of personal health data and images~\citep{HIPAA}.
Thus, directly sharing and pooling medical images across different sites/centers is typically infeasible in real-world practice.

\begin{figure*}[t]%[!tbp]
\setlength{\belowcaptionskip}{-2pt}
\setlength{\abovecaptionskip}{0pt}
\setlength{\abovedisplayskip}{0pt}
\setlength{\belowdisplayskip}{0pt}
\center
 \includegraphics[width= 0.9\linewidth]{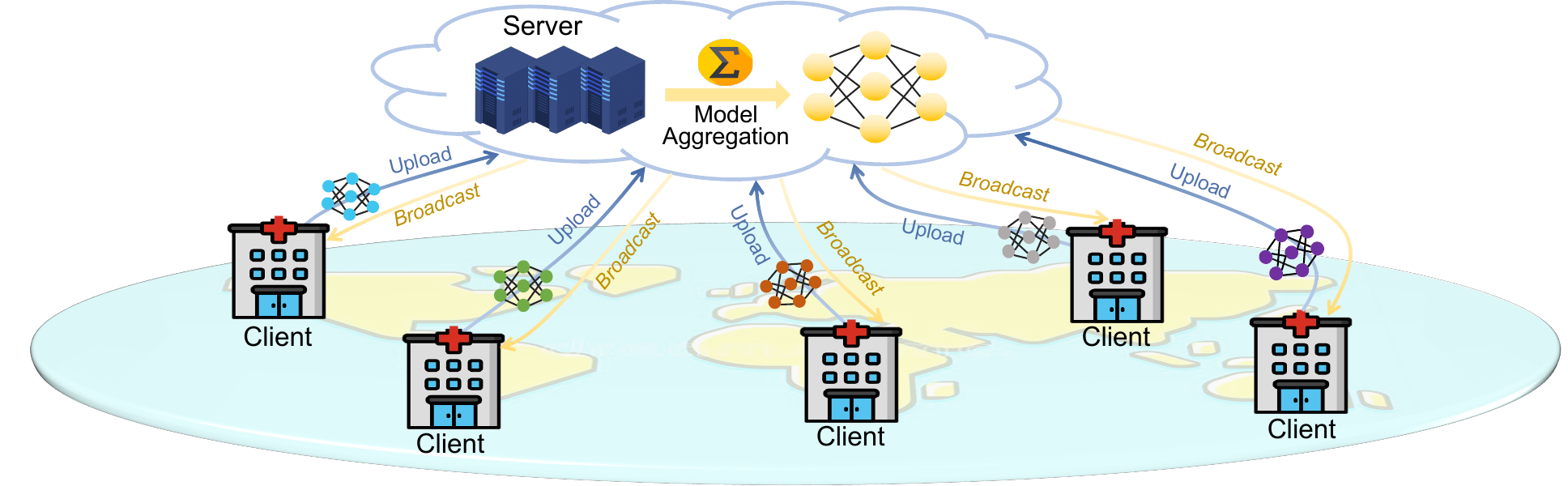}%{FL_v1.pdf}%
% \vspace{2}
 \caption{Overview of federated learning (FL) for medical image analysis, including a server and multiple clients. 
Each selected client trains a model on its local dataset. The server collects the local models and calculates a global model that is broadcast to all the selected clients for deployment.}
 \label{fig_FL_overview}
\end{figure*}

As a promising solution for dealing with the small-sample-size problem and protecting individual privacy, 
federated learning~\citep{FL,FL3,FL2} has become a spotlight research topic in recent years, which aims to train machine learning models in a collaborative manner without exchanging/sharing data among different sites.
As an emerging machine learning paradigm, federated learning deliberately avoids demand for all the medical data residing in one single site. 
Instead, as shown in Fig.~\ref{fig_FL_overview}, it depends on model aggregation/fusion techniques to jointly train a global model which is then sent/broadcast to each site for fine-tuning and deployment.

\subsection{Related Surveys}
There have been several survey papers on federated learning~\citep{li2021survey,li2020federated,yang2019federated,rahman2021challenges,zhang2021survey,yin2021comprehensive}, but
further technical details about facilitating federated learning in medicine and healthcare are not yet covered. 
Several recent surveys introduce the applications of federated learning in medicine and healthcare areas~\citep{FL-Medical1,rajendran2021cloud,nguyen2022federated,pfitzner2021federated,rieke2020future}.   
However, some of them focus on electronic health records~\citep{FL-Medical1,rajendran2021cloud} or internet of medical things~\citep{aouedi2022handling}, without paying attention to medical imaging. And some survey papers cover very broad areas in medicine and healthcare applications~\citep{nguyen2022federated,pfitzner2021federated}, without detailed introduction on federated learning in medical image analysis. 
%%%%%%%%%%%%%%%%%%%%%%%%%%%%%%%%%%%%%%%%% 
%%%%%%%%%%%%%%%%%%%%%%%%%%%%%%%%%%%%%%%%%%%%%%%%%%%%%%%%%%%%%%%%%%%%%%%%%%%%%%
A recent survey also reviews the application of federated learning on medical image analysis~\citep{sohan2023systematic}.
Our survey paper reviews and discusses the most recent advances in federated learning for medical image analysis and has significant differences from the previous one in the following aspects.

\begin{itemize}

\item
\textbf{Different Coverage.}
The previous review was limited over the time period before December 2022.
Our paper covers the papers from 1 January 2017 to 31 October 2023.
With a broader coverage, most recent advances of the state-of-the-art models or methods in federated learning in medical image analysis (\eg, transformer) have been included in our paper.

\item
\textbf{Software Platforms.}
The previous survey does not include federated learning software platforms that have been applied to medical image analysis.
Our paper emphasizes the implementation of federated learning techniques for medical image analysis. 
Specifically, we introduce some new and influential FL software platforms and benchmark medical imaging datasets for federated learning research in medical imaging.  

\item 
\textbf{Experimental Study.}
The previous survey only conducts a survey and summary of published papers.
As for our work, besides a summary of existing work, we also conduct an experimental study to evaluate typical federated learning methods for
medical image analysis empirically. This could offer the readers a more intuitive understanding of this research topic.

\item
\textbf{Future Direction and New Arisen Problems.}
Due to the inclusion of the most recent papers, our survey paper offers a more comprehensive summary of newly arisen research problems (\eg, model generalizability for unseen clients, and FL for medical video analysis) and points out a broader range of future directions of federated learning for medical image analysis.

\item
\textbf{Different Perspective and Organization.}
Different from previous surveys that are based on multiple research issues in federated learning, we summarize the existing methods from a system perspective.
Specifically, we categorize different approaches into three groups: 1) client-end learning methods, 2) server-end learning methods, and 3) server-client communication methods. This categorization can be more intuitive and clear to picture federated learning.
When elaborating on the methods in each group, we have designed a novel ``question-answer'' paradigm to introduce the motivation and mechanism of each method. 
We deliberately extract the common questions behind different methods and pose them first in each subsection.
These questions stem from the characteristics of medical imaging, thus it helps provide more insights into different methods.

\end{itemize}

\if false
\emph{\textbf{First}},
we summarize the existing methods from a system perspective.
Specifically, we categorize different approaches into three groups: 1) client-end learning methods, 2) server-end learning methods, and 3) server-client communication methods.
Different from previous surveys that are based on multiple research issues in federated learning, this categorization can be more intuitive and clear to picture federated learning.
\emph{\textbf{Second}},
when elaborating on the methods in each group, we have designed a novel ``question-answer'' paradigm to introduce the motivation and mechanism of each method. 
We deliberately extract the common questions behind different methods and pose them first in each subsection.
These questions stem from the characteristics of medical imaging, thus this ``question-oriented" approach of introduction is helpful for providing more insights into different methods.
\emph{\textbf{Third}}, 
we emphasize the implementation of federated learning techniques for medical image analysis. 
Specifically, we introduce popular software platforms and benchmark medical imaging datasets for federated learning research in medical imaging.  
\emph{\textbf{In addition}}, we also conduct an experiment on a benchmark medical image dataset to illustrate the utility and effectiveness of several typical federated learning methods.
\fi
%%%%%%%%%%%%%%%%%%%%%%%%%%%%%%%%%%%%%%%%%%%%%%%%%%%%%%%%%%%%%%%%%%%%%%%%%%%%%%
%%%%%%%%%%%%%%%%%%%%%%%%%%%%%%%%%%%%%%%%% 
\begin{figure*}[t]%[!tbp]
\setlength{\belowcaptionskip}{-2pt}
\setlength{\abovecaptionskip}{0pt}
\setlength{\abovedisplayskip}{0pt}
\setlength{\belowdisplayskip}{0pt}
\center
 \includegraphics[width= 0.5\linewidth]{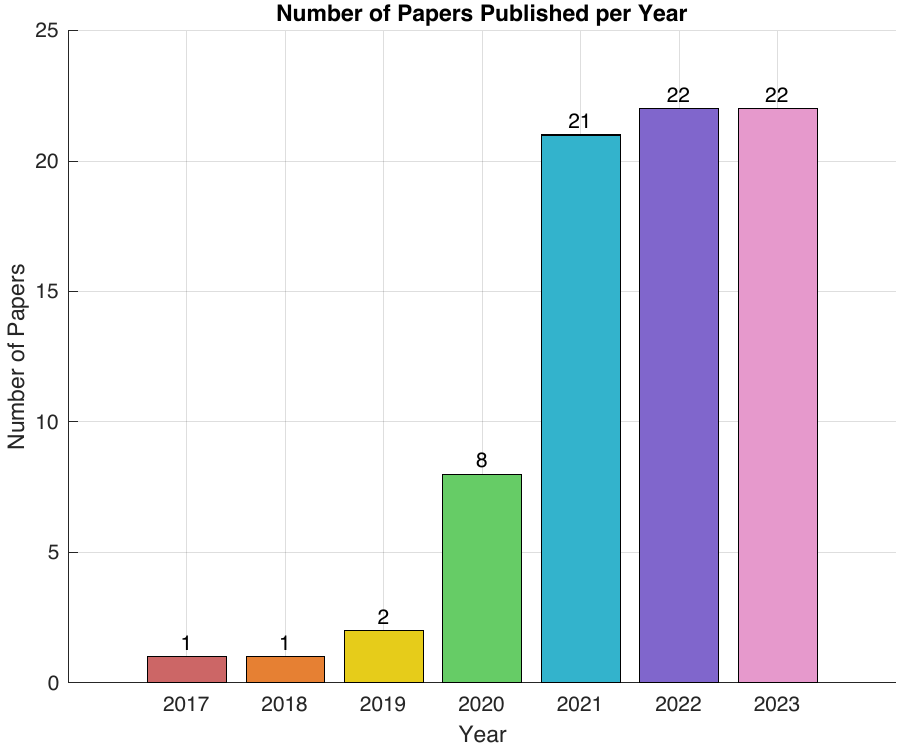}%{FL_v1.pdf}%
% \vspace{2}
 \caption{
Overview of the number of papers (in terms of published years) that have been collected for this survey on federated learning in medical image analysis.}
 \label{fig_papers_year}
\end{figure*}
%%%%%%%%%%%%%%%%%%%%%%%%%%%%%%%%%%%%%%%%%%%%%%%%%%%%%%%%%%%%%%%%%%%%%%%%%%%%%%

\subsection{Searching and Analysis Process}
The first paper on federated learning was released in the year of 2016, thus 
the searching process for this survey ranges from 1st January 2017 to 31st October 2023 (see Fig.~\ref{fig_papers_year}).
There are three steps in conducting this survey paper.
\emph{\textbf{First}}, we performed a literature search using the academic databases and engines, including 1) IEEE Xplore, 2) ACM Digital Library, 3) Science Direct, 4) Springer Link, 5) Web of Science, 6) Google Scholar and 7) PubMed.  
\emph{\textbf{Second}}, we refined the initial result from the digital libraries by removing duplicated papers and papers that do not have close relationship with medical imaging (\eg, non-imaging healthcare data).
\emph{\textbf{Third}}, we analyzed the refined papers, extracted the common research questions and technical solutions, and constructed this survey paper. 

The remainder of this paper is organized as follows. 
In Section~\ref{Background}, we introduce the background and motivation of federated learning.
We summarize existing federated learning studies for medical image analysis in Section~\ref{Methods}.
In Section~\ref{Software}, software platforms that support federated learning system development are presented.
In Section~\ref{Datasets}, we introduce medical image datasets that have been widely used in federated learning research.
We conduct an experimental study in Section~\ref{Experiment} to compare several federated learning methods.
Challenges and potential research opportunities are discussed in Section~\ref{Discussions}. 
Finally, we conclude this survey paper in Section~\ref{Conclusion}.

%%%%%%%%%%%%%%%%%%%%%%%%%%%%%%%%%%%%%%%%%%%%%%%%%%%%%%%%%%%%%%%%%%
\section{Background}  \label{Background}
\subsection{Motivation}
\subsubsection{Privacy Protection in Medical Image Analysis}
Patient data protection has become an important issue in the digital era.
Using and selling patient data has many negative implications~\citep{chiruvella2021ethical}.
Thus, many governments have introduced tough new laws and regulations on privacy data protection, such as the CCPA in the United States~\citep{CCPA} and GDPR in Europe~\citep{GDPR}.
Collecting, sharing, and processing of personal data are strictly constrained, and violating these laws and regulations may face high-cost penalties~\citep{satariano2019google}. 
With these strict restrictions from laws, medical images, one of the most important privacy information, cannot be easily shared among different sites/centers.
To this end, federated learning, a distribution-oriented machine learning paradigm without cross-site data sharing, has emerged as a promising technique for developing privacy-preservation machine learning models, thus paving the way for the applications of medical artificial intelligence in real-world practice.

\subsubsection{Challenges of Medical Image Analysis}
The conventional approach to training machine learning models in medical image analysis involves utilizing data from a single site or center. 
However, this method is usually subject to limited sample size. 
It is common that there are very limited number of images in local datasets. 
This situation often arises due to the high costs associated with imaging and labeling procedures. Consequently, the datasets suffer from the ``small-sample-size" problem~\citep{raudys1991small,vabalas2019machine}. This issue can severely impact the learning performance of machine learning models, leading to suboptimal results that lack statistical significance.
%2) \textbf{Data Distribution Bias.} 
Another significant concern is the inherent bias in the distribution of data collected from a specific site or center. 
These datasets may not accurately represent the true data distribution, thereby introducing bias into machine learning models. 
For instance, it is common to encounter unbalanced data in medical sites, where the number of healthy subjects significantly outweighs that of patients. Such imbalances can skew the model's predictions and compromise its effectiveness in real-world applications. 
%3) \textbf{Limited Generalizability.} 
In addition, medical image datasets collected from a specific site often reflect the characteristics and demographics of the local patient population. Consequently, models trained solely on such data may fail to capture the variability present in broader patient cohorts or diverse clinical settings.
\if false
The traditional way to train machine learning models is to use medical images from a specific site/center. 
It has at least two following drawbacks.
\begin{enumerate}
\item[(1)] 
Due to the cost of imaging and labeling, the amount of images in local datasets is usually small. 
This is the well-known ``small-sample-size" problem~\citep{raudys1991small,vabalas2019machine}. 
This problem may lead to sub-par learning performance of a machine learning model, and produce results that lack statistical significance. 

\item[(2)]
Data from a specific site/center may be biased in distribution and not representative of the true data distribution. 
For instance, it is not unusual that medical sites contain unbalanced data, where healthy subjects significantly outnumber the patients.
\end{enumerate}
\fi
Federated learning helps address these limitations, aiming to ``pool" medical images together in a distributed way, thereby greatly increasing the sample size. 
This can effectively take advantage of available data from multiple sites to enhance statistical power of machine learning models.

%%%%%%%%%%%%%%%%%%%%%%%%%%%%%%%%%%%%%%%%%%%%%%%%%%%%%%%%%%%%%%%%%%%%%%%%%%%%%%%%
\subsection{Problem Formulation of Federated Learning}
Suppose there are $N$ independent clients (sites) with their own datasets \{$\mathcal{D}_1, \mathcal{D}_2, \cdots, \mathcal{D}_N$\}, respectively.
Each of the clients (sites) cannot get access to others' datasets. % besides its own. % for data augmentation.
Federated learning (FL) aims to collaboratively train a machine learning model $\mathcal{M}^{*}$ by gathering information from those $N$ clients (sites) without exchanging/sharing their raw data.
The ultimate output of FL is the learned model $\mathcal{M}^{*}$ which is broadcast to each client for deployment, 
and the generalizability of $\mathcal{M}^{*}$ by FL should outperform each local model $\mathcal{M}_{i}$ (typically with the same model architecture as $\mathcal{M}^{*}$) learned through local training.
%%%%%%%%%%%%%%%%%%%%%%%%%%%%%%%%%%%%%%%%%%%%%%%%%%%%%%%%%%%%%%
\subsection{Typical Process of Federated Learning}
In Fig.~\ref{fig_FL_overview}, we illustrate the typical process of federated learning that is embodied in a ``client-server" architecture. 
This process 
encompasses the \emph{Federated Averaging} (FedAvg) algorithm proposed by McMahan~\etal\citep{FL}.
It serves as the foundation of most popular algorithms for federated learning. 
A server in a federation triggers and orchestrates the entire training process (without accessing clients' private data) until a certain stop criterion is met.
We summarize a typical workflow of federated learning as follows.

\begin{enumerate}
\item[1)] \textbf{Client Selection and Initialization.}
The server selects a set of clients that meet certain requirements. 
For example, a medical site/center might only check in to the server when it can correctly get access to the intranet of a federation with relatively good bandwidth. 
Some recent FL models dynamically select clients that meet certain requirements such as training efficiency~\citep{zhang2021dynamic} or anomaly score~\citep{MICCAI-10}.
A global model is initialized on the central server. This model serves as the starting point for training across different medical sites/centers (\ie, clients).

\item[2)] \textbf{Local Training.}
The global model is sent to all the participating medical sites/centers. 
Each site/center trains a machine learning model (\eg, U-Net) on its local medical imaging data using certain optimization methods (\eg, stochastic gradient descent).
With the development of artificial intelligence, some recent work introduced more advanced models for client training such as vision transformer~\citep{yan2023label}. 
Since the data never leaves its original location, this process can enhance privacy and security.

\item[3)] \textbf{Model Upload.}
After local training, each medical site/center calculates the updates to the model (\eg, gradients or model changes) and sends/uploads these updates back to the central server. Importantly, only model updates are shared, not the data itself.

\item[4)] \textbf{Aggregation.}
The central server aggregates all the updates uploaded by the clients, typically using certain algorithms that ensure a fair and effective combination of the different contributions. This aggregation results in an updated global model.
While classic FL systems use equal weights for aggression, some recent models explore using more adaptive weighting strategies~\citep{li2022integrated} to enhance training efficiency. 

\item[5)] \textbf{Broadcast.}
During the broadcasting step, the server sends the updated model parameters or gradients to the clients, enabling them to perform local computations and contribute to the collaborative model training process. By efficiently distributing model updates, the broadcasting step facilitates synchronized model updates across the federated network while minimizing communication overhead.
Research on this topic has focused on optimizing communication protocols and minimizing communication overhead~\citep{yang2021flop,zhang2021dynamic} while ensuring efficient dissemination of model updates.

\item[6)] \textbf{Iteration and Convergence.}
The above steps are repeated for several iterations. With each round, the global model becomes more refined and accurate, as it learns from a diverse set of data sources.
This process continues until the model reaches a satisfactory level of accuracy or a predefined number of iterations are completed.
Recent research work focuses on improving the overall training efficiency and accelerate the convergence~\citep{ISBI_FL1}. 

\item[7)] \textbf{Deployment.}
The final global model is then deployed for use in applications, maintaining the benefits of having learned from a wide and diverse set of data sources.
In real-world practice, several factors or challenges need to be considered such as compatibility with
existing hospital systems, integration challenges, and user adoption hurdles. 

\end{enumerate}

%%%%%%%%%%%%%%%%%%%%%%%%%%%%%%%%%%%%%%%%%%%%%%%%%%%%%%%%%%%%%%%%%%%%%%%%%%%%%%%%%%%%%%%%%%%%%%%%%%%%%%%%%%%%%%%%%%%%%%%%%%%%%%%%%%%%%%%%%%%%%%%%%%%%%%%%%%%%%%%%%

%%%%%%%%%%%%%%%%%%%%%%%%%%%%%%%%%%%%%%%% 
\begin{figure*}[t]%[!tbp]
\setlength{\belowcaptionskip}{0pt}
\setlength{\abovecaptionskip}{0pt}
\setlength{\abovedisplayskip}{0pt}
\setlength{\belowdisplayskip}{0pt}
\center
 \includegraphics[width= 1.0\linewidth]{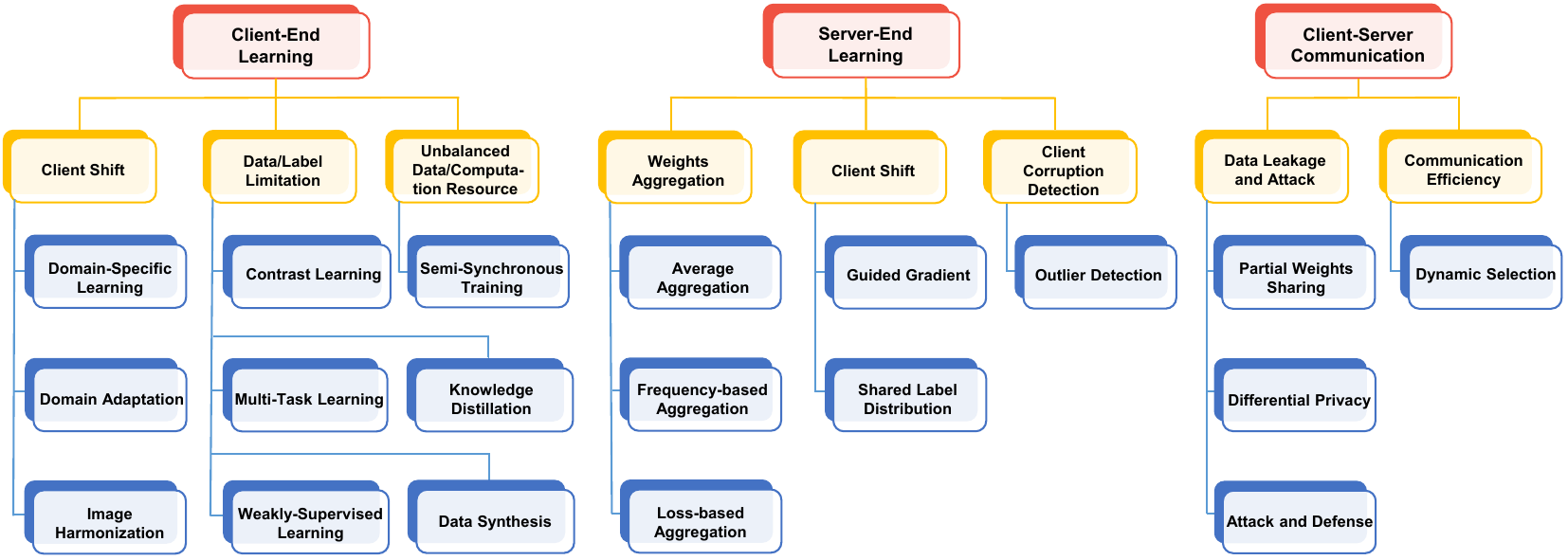}
% \vspace{2}
 \caption{Overview of federated learning (FL) methods for medical image analysis.} 
 \label{fig_Taxonomy}
\end{figure*}
%%%%%%%%%%%%%%%%%%%%%%%%%%%%%%%%%%%%%%%%%%%%%%%%%%%%%%%%%%%%%%%%%%%%%%%%%%%%%%%%%
%%%%%%%%%%%%%%%%%%%%%%%%%%%%%%%%%%%%%%%%%%%%%%%%%%%%%%%%%%%%%%%%%%%%%%%%%%%%%%%%%
\subsection{Types of Federated Learning}
\subsubsection{Horizontal Federated Learning}
Horizontal federated learning~\citep{yang2019federated}, also known as homogeneous federated learning, is characterized by data distribution across different entities that share the same feature space but have different samples. In the context of medical image analysis, this can be thought of as different medical institutions holding medical imaging data (\eg, MRIs, X-rays) of different patients.

\textbf{Examples in Medical Image Analysis.}
Consider multiple hospitals across different regions participating in a study to improve the diagnosis of lung diseases using chest X-rays. Each hospital has its own set of patient data (images), but the features extracted from these images (\eg, lung size, shape, and texture) are similar across all datasets. Horizontal FL allows these hospitals to collaboratively train a model to diagnose lung diseases more accurately without sharing the actual patient data.
%%%%%%%%%%%%%%%%%%%%%%%%%%%%%%%%%%%%%%%%%%%%%%%%%%%%%%%%%%%%%%%%%%%%%%%%%%%%%%%%%
\subsubsection{Vertical Federated Learning}
Vertical federated learning~\citep{yang2019federated}, or heterogeneous federated learning, occurs when different entities possess different feature sets for the same samples. In medical imaging, this translates to different institutions having different types of data (\eg, omics data, demographic information, and imaging data) for the same set of patients.

\textbf{Examples in Medical Image Analysis.}
Vertical FL is increasingly prevalent in medical imaging studies due to the multidisciplinary nature of healthcare data. For instance, a hospital might have imaging data, while a research lab could hold genomic data for the same set of patients. Through vertical FL, these diverse datasets can be utilized to create more comprehensive models for disease diagnosis and prognosis, without compromising patient privacy.
%%%%%%%%%%%%%%%%%%%%%%%%%%%%%%%%%%%%%%%%%%%%%%%%%%%%%%%%%%%%%%%%%%%%%%%%%%%%%%%%%
%%%%%%%%%%%%%%%%%%%%%%%%%%%%%%%%%%%%%%%%%%%%%%%%%%%%%%%%%%%%%%%%%%%%%%%%%%%%%%%%%
%%%%%%%%%%%%%%%%%%%%%%%%%%%%%%%%%%%%%%%%%%%%%%%%%%%%%%%%%%%%%%%%%%%%%%%%%%%%%%%%%
\section{Federated Learning for Medical Image Analysis} 
\label{Methods}
%%%%%%%%%%%%%%%%%%%%%%%%%%%%%%%%%%%%%%%%%%%%%%%%%%%%%%%%%%%%%%%%%%%%
\subsection{Methods Overview: A System Perspective}
Federated learning (FL)
provides a generic framework for distributed learning with privacy preservation.
Most existing machine/deep learning methods can be plugged and integrated into an FL framework.
For example, a U-Net~\citep{U-net} can be used in each client for medical image segmentation and is trained in a federated manner.
Federated learning is concerned with multiple issues such as data, machine learning models, privacy protection mechanisms, and communication architecture.
As shown in Fig~\ref{fig_Taxonomy}, from a system perspective, we categorize existing FL approaches for medical image analysis into three groups: 
1) client-end methods, 2) server-end methods, and 3) communication methods. 
In each group, different methods are clustered according to the specific research problems they aim to address which will be elaborated in the following sections.  

%%%%%%%%%%%%%%%%%%%%%%%%%%%%%%%%%%%%%%%%%%%%%%%%%%%%%%%%%%%%%%%%%%%%%%%%%%%%%%%%%

\subsection{Client-End Learning}\label{Methods-B}
%%%%%%%%%%%%%%%%%%%%%%%%%%%%%%%%%%%%%%%%%%%%%%%%%%%%%%%%%%%%%%%%%%%%%%%%%%%%%%%%%%%%%%
\subsubsection{Client End: Domain Shift Among Clients}
\vspace{3pt}

\noindent  \textbf{Problem:} \emph{
This research addresses the challenge of significant cross-site data distribution variance in medical imaging, often resulting from varying scanning settings and diverse subject populations across different sites. The focus is on developing strategies to mitigate this variance's adverse impact on the accuracy and reliability of FL model training.
}

\begin{figure}[t]%[!tbp]
\setlength{\belowcaptionskip}{0pt}
\setlength{\abovecaptionskip}{0pt}
\setlength{\abovedisplayskip}{0pt}
\setlength{\belowdisplayskip}{0pt}
\centering
 \includegraphics[width= 0.66\linewidth]{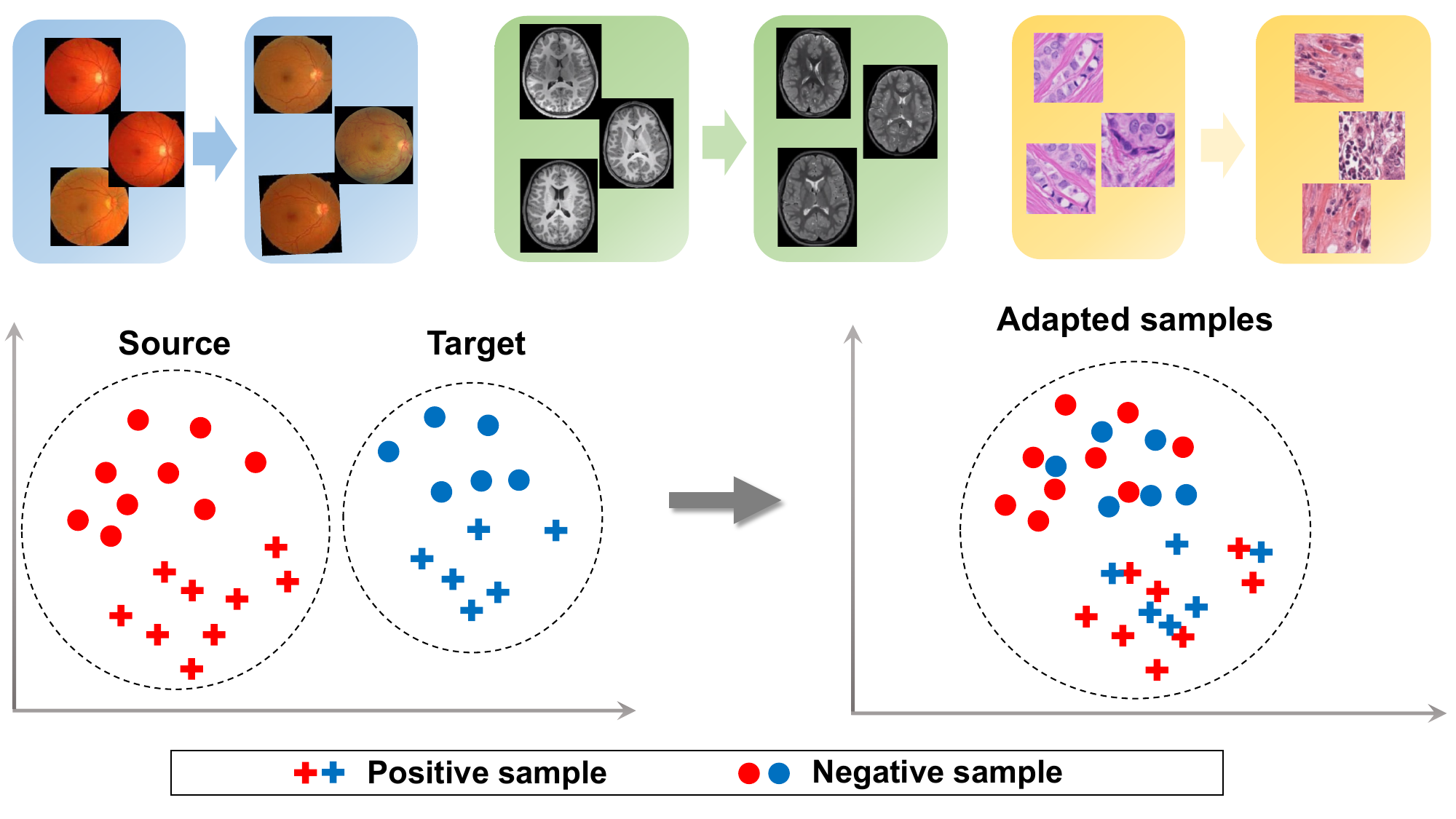}
% \vspace{2}
 \caption{Domain shift among different medical sites (domains). Domain adaptation aims to reduce domain differences and enhance machine learning performance across different sites. Image courtesy to Guan~\etal~\citep{guan2023domainatm}.}
 \label{fig_DS}
\end{figure}

\vspace{3pt}
In practice, multi-site medical images may have significantly different data distributions, which is the well-known ``domain shift" problem~\citep{guan2022domain} (also referred to as ``client shift" in an FL system). 
As shown in Fig.~\ref{fig_DS}, images from different imaging sites have significantly different intensity distributions. When projected in the feature space, the domain shift may negatively influence machine learning performance. Thus certain techniques, \eg, domain adaptation, are adopted to alleviate this issue by making the distribution differences smaller. 
In an FL system, domain shifts may cause difficult convergence of the global model and performance degradation of some clients. 
In the following, we present the relevant studies that focus on reducing domain shift among clients for FL research.
%%%%%%%%%%%%%%%%%%%%%%%%%%%%%%%%%%%%%%%%%%%%%%%%%%%%%%%%%%%%%%%%%%%%%%%
%%%%%%%%%%%%%%%%%%%%%%%%%%%%%%%%%%%%%%%% 
%%%%%%%%%%%%%%%%%%%%%%%%%%%%%%%%%%%%%%%%%%%%%%%%%%%%%%%%%%%%%%%%%%%%%%%%

\vspace{3pt}
\noindent \textbf{(1) Domain-Specific Learning.} 
This method uses client data to locally fine-tune the global model and alleviate negative influence of client shift.
%Federated learning aims to train a global model that fits well with all clients. 
%Due to cross-site data heterogeneity, the global model may not be able to achieve good performance for all clients.
%One strategy is fine-tuning the global model using domain-specific (local) data to make it more suitable for a specific client.
This method is also known as customized/personalized FL~\citep{wicaksana2022customized,personalized1}.
Feng~\etal\citep{feng2022specificity} propose an encoder-decoder structure within an FL framework for magnetic resonance (MR) image reconstruction. 
A shared encoder is maintained on the server end to learn domain-invariant representations, while a client-specific decoder is trained with local data to take advantage of domain-specific properties of each client.
Similar strategies can also be found in \citep{zhang2022splitavg,wicaksana2022customized}. 
Chakravarty~\etal\citep{ISBI_FL2} propose a framework that combines a Convolutional Neural Network (CNN) and a Graph Neural Network (GNN) to tackle the domain shift problem among clients and apply it to chest X-ray image classification.
Specifically, model weights of the CNN  are shared across clients to learn site-independent features.
To address site-specific data variations, a local GNN is built and fine-tuned with local data in each client for disease classification. 
In this way, both site-independent and site-specific features can be learned.  
Xu~\etal\citep{xu2022closing} propose an ensemble-based framework to deal with the client shift for medical image segmentation.
Their framework is composed of a global model, personalized models, and a model selector.
Instead of only using the global model to fit all the client data, they propose to leverage all the produced personalized models to fit different client data distributions through a model selector.
Jiang~\etal\citep{jiang2023iop} propose to train a locally adapted model that accumulates both global gradients (aggregated from all clients) and local gradients (learned from local data) to optimize the model performance on each client. 
This helps effectively avoid biased performance of the global model on different clients caused by client domain shift. 
Ke~\etal\citep{ISBI_FL3} build an FL framework based on a Generative Adversarial Network (GAN) to facilitate harmonization (color normalization) of histopathological images.
In this method, each client trains a local discriminator to capture client-specific image style, while the server maintains and updates a global generator model to generate domain-invariant images, thus achieving histopathological image harmonization.
Similarly, Wagner~\etal\citep{MICCAI-13} propose a GAN model for histopathological image harmonization. 
In their method, a reference dataset is assumed to be accessible for all clients, which can help the training of all the local GANs at each client.

\vspace{3pt}
\noindent \textbf{(2) Domain Adaptation.} 
This method uses domain adaptation to reduce medical data distribution differences of clients.
Domain adaptation is a sound machine learning technique that has been widely used in medical image analysis~\citep{guan2022domain}. It aims to reduce domain shift (in terms of certain distances) among different medical image datasets and enhance the generalizability of a machine learning model. 
Many medical-related FL studies resort to domain adaptation for improved performance. 
Li~\etal\citep{li2020multi} use domain adaptation to align distribution differences of functional MRI data among clients. In their method, data in each client are added with noise to enhance privacy protection.
A domain discriminator/classifier is trained on these data with noises to reduce domain shift.
Dinsdale~\etal\citep{MICCAI-12} propose a domain adaptation-based FL framework to remove domain shift among clients caused by different scanners. In their framework, medical image features are assumed to follow Gaussian distributions, and the mean and standard deviation of the learned features can be shared among clients. During the training of each client model, a label classifier and a domain discriminator are jointly trained to learn features that are domain-invariant, \ie, removing domain shift. 
Andreux~\etal\citep{andreux2020siloed} leverages batch normalization (BN) in a deep neural network to handle client (histopathology datasets) shift.
Guo~\etal\citep{guo2021multi} propose a federated learning method for MRI reconstruction, where the learned intermediate latent features among different
clients are aligned with the distribution of latent features of a reference site.
%%%%%%%%%%%%%%%%%%%%%%%%%%%%%%%%%%%%%%%%%%%%%%%%%%%%%%%%%%%%%%%%%%%%%%%%%%%%%%%%%%%%

\vspace{3pt}
\noindent \textbf{(3) Image Harmonization.} 
This method typically uses image-to-image translation models to harmonize the medical images of different clients.
After harmonization, all the medical images are expected to have similar styles, thus reducing domain shift.
Qu~\etal\citep{qu2022handling} propose a generative replay strategy to handle data heterogeneity among clients. 
They first train an auxiliary variational autoencoder (VAE) to generate medical images that resemble the input images. 
Then each client can optimize their local classifier using both the real local data and synthesized data with similar data distribution of other clients. In this way, domain shift can be reduced.
Yan~\etal\citep{yan2020variation} employ cycleGAN~\citep{cycleGAN} to minimize the variations of medical images among clients. One client/site with low data complexity is selected as a reference, and then cycleGAN is used to harmonize medical images from other clients to the reference site.
Jiang~\etal\citep{AAAI-2} propose a frequency-based harmonization method to reduce client shift in medical images.
Medical images are firstly transformed into frequency domain and phase components are kept locally, while the average amplitudes from each client are shared and then normalized to harmonize all the client medical images.

%%%%%%%%%%%%%%%%%%%%%%%%%%%%%%%%%%%%%%%%%%%%%%%%%%%%%%%%%%%%%%%%
\subsubsection{Client End: Limited Data and Labels}

\vspace{3pt}

\noindent  \textbf{Problem:} \emph{
%Medical imaging datasets are often small-sized and lack label/annotation information, so how to avoid their negative influence on model training (\eg, biased training)?
This research tackles the prevalent issue in medical imaging where datasets are frequently small-sized and deficient in label information. The focus is on developing strategies to mitigate their negative impact on FL model training and prevent biased results.
}

\vspace{3pt}

In real-world practice, there are often limited medical images in one client/site, and labeled medical images are even fewer due to the high cost of image annotation/labeling. A client model may be badly trained with limited labeled data, which can cause negative influences on the entire federation. Therefore, how to alleviate the small-sample-size problem is an important topic of FL in medical image analysis. 

\vspace{3pt}
\noindent \textbf{(1) Contrast Learning.}  
Contrastive learning~\citep{chaitanya2020contrastive,he2020momentum,misra2020self} is a self-supervised learning method where models learn to distinguish between similar and dissimilar data points.
%Contrast learning can learn useful representations of images by using unlabeled data and it is particularly effective in image recognition where learning to identify subtle differences or similarities can significantly enhance performance.
A model trained with contrast learning can provide good initialization for further fine-tuning (with a few labeled data) on downstream tasks. 
%Contrast learning has been introduced into federated learning for handling medical data shortage. 
\citep{wu2022distributed,MICCAI-3} use contrast learning to pretrain the encoder of a U-Net in each client, then the global U-Net is fine-tuned with limited labeled medical images. In this way, the negative influence caused by the shortage of labeled medical images can be largely reduced. Similar strategies can be found in~\citep{MICCAI-2}.

\vspace{3pt}
\noindent \textbf{(2) Multi-Task Learning.}
Multi-task learning~\citep{multi-task1} is an effective learning paradigm for data augmentation.
%It typically solves multiple but related learning tasks jointly, which can exploit commonalities across tasks.
%When the training data for each task are small-sized, jointly learning of different tasks can share data which is an effective approach for data augmentation. 
Smith~\etal\citep{multi-task2} propose a novel optimization framework, \ie, MOCHA, which extends classic multi-task learning in the federated environment. 
Huang~\etal\citep{huang2022federated} build a federated multi-task framework in which several related tasks, \ie, attention-deficit/hyperactivity disorder (ADHD), autism spectrum disorder (ASD), and schizophrenia (SCZ), are jointly trained. 
In this method, encoders for each task in clients are federated to derive a global encoder that can learn common knowledge among related mental disorders.

\vspace{3pt}
\noindent \textbf{(3) Weakly-Supervised Learning.} 
Weakly-supervised learning~\citep{zhou2018brief} is an extensive group of methods that train a model under weak supervision, including 1) Incomplete supervision~\citep{SSL-survey,van2020survey}; 2) Inexact supervision~\citep{quellec2017multiple,carbonneau2018multiple}; and 3) Inaccurate supervision~\citep{song2022learning,frenay2013classification}. 
\if false
Weak supervision information typically includes three types. 
1) \textit{Incomplete supervision}. Only
a small subset of labeled training data is provided
while the other data has no labels. Semi-supervised learning~\citep{SSL-survey,van2020survey} is a popular solution for such scenarios.
2) \textit{Inexact supervision}. Only coarse-grained labels are provided for the training data. 
Multiple-instance learning~\citep{quellec2017multiple,carbonneau2018multiple} is a representative method to handle this problem.
3) \textit{Inaccurate supervision}. Not all the provided labels are correct.
Learning from noisy labels~\citep{song2022learning,frenay2013classification} is the corresponding technique. 
\fi
Yang~\etal\citep{yang2021federated} introduce semi-supervised learning into FL for chest CT segmentation. In their method, unlabeled CT images are leveraged to assist the federated training.
For unlabeled CT images in a client, the global model assigns them pseudo labels. Meanwhile, it also outputs predictions on augmented data of the original unlabeled images. 
A consistency loss is utilized on these predictions to further adjust the global model weights.
Lu~\etal\citep{lu2022federated} use multiple-instance learning for local model training on the task of pathology image classification.
Whole slide images (WSIs) and weak annotation (\eg, patient or not) are used as the input, with no region-based labels provided.   
And multiple patches (instances) of a WSI are fed into a network for training.
Kassem~\etal\citep{kassem2022federated} build a semi-supervised FL system for surgical phase recognition based on laparoscopic
cholecystectomy videos.
The key idea is to leverage the temporal information in labeled videos to guide unsupervised learning on unlabeled videos.

\vspace{3pt}
\noindent \textbf{(4) Knowledge Distillation.} 
Knowledge distillation~\citep{hinton2015distilling} is a process where a smaller, more efficient model (the ``student") is trained to replicate the behavior of a larger, more complex model (the ``teacher"). 
This is achieved by using the outputs of the teacher model as a guide for training the student model, effectively transferring the knowledge. 
%For example, a large neural network trained for medical image classification can be used to train a smaller network, enabling it to achieve similar accuracy can significantly reduce the demand for labeled data and computational resources.
Kumar~\etal\citep{kumar2021medisecfed} leverage knowledge distillation for COVID-19 detection in chest X-ray images.
The network trained on similar data (other chest X-ray image datasets) is used as a ``teacher", while the client model is a ``student". 
By matching the softmax activation output of the teacher, the student (client model) can learn useful knowledge for the task. 
In this way, it alleviates the demand for large data during the FL training process.
He~\etal\citep{He2023ISBI} use knowledge distillation to address the problem of weakly-supervised learning in heterogeneous 3D MR knee images.
Unlabeled data in the client is used to distill knowledge from the large-scale national data repository to improve the performance of the collaborative model.

\vspace{3pt}
\noindent \textbf{(5) Data Synthesis.}
This method typically uses generative models (\eg, GAN) to create synthesized medical images as data enhancement for federated learning.
Zhu~\etal\citep{MICCAI-11} propose an FL framework with virtual sample synthesis for medical image analysis. 
Given an image $\mathbf{x}$ in the client, the authors first use Virtual Adversarial Training~\citep{VAT} to generate synthetic images that are similar to $\mathbf{x}$, and then use all the synthesized images for local model training.
Chang~\etal\citep{chang2023mining} present a novel GAN-based synthetic learning approach for extracting information from each client to generate a homogeneous dataset with entirely synthetic medical images for downstream applications.
Peng~\etal\citep{peng2022fedni} propose a federated graph learning framework for brain disease prediction, where a Graph Convolutional Network (GCN) is used as the learning model in each client. 
Considering the missing nodes and edges when separating the global graph into local graphs, the authors leverage network inpainting to predict the missing nodes and their associated edges. 
This helps complete the graphs for GCN training in each client, with results suggesting its effectiveness in graph data synthesis and augmentation. 

%%%%%%%%%%%%%%%%%%%%%%%%%%%%%%%%%%%%%%%%%%%%%%%%%%%%%%%%%%%%%%%%%%
%%%%%%%%%%%%%%%%%%%%%%%%%%%%%%%%%%%%%%%% 
\begin{figure}[!tbp]
\setlength{\belowcaptionskip}{0pt}
\setlength{\abovecaptionskip}{0pt}
\setlength{\abovedisplayskip}{0pt}
\setlength{\belowdisplayskip}{0pt}
\center
 \includegraphics[width= 0.56\linewidth]{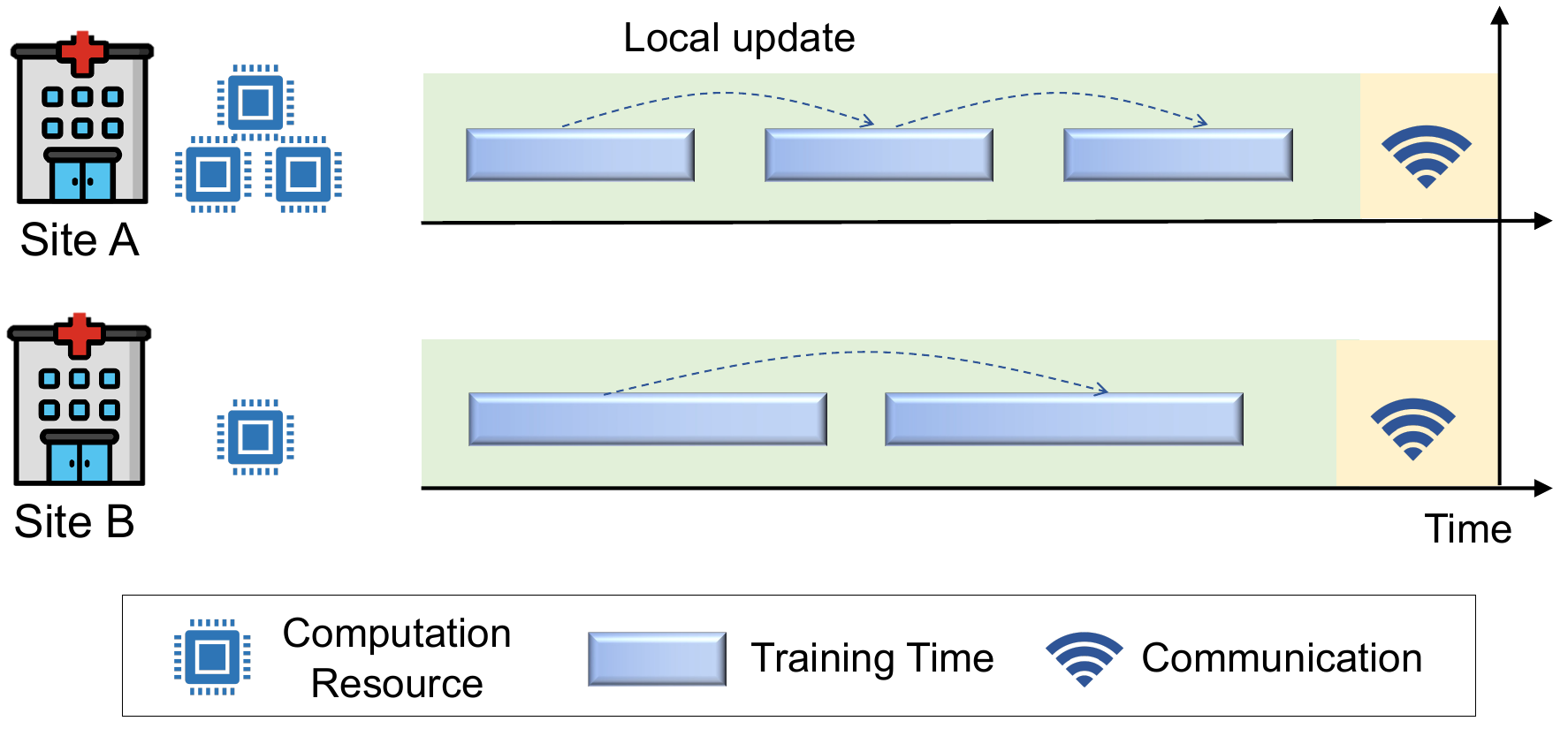}
% \vspace{2}
 \caption{Different local updates for clients with different computation and data resources.}
 \label{fig_unbalance}
\end{figure}
%%%%%%%%%%%%%%%%%%%%%%%%%%%%%%%%%%%%%%%%%%%%%%%%%%%%%%%%%%%%%%%%%%%%%%%%

\subsubsection{Client End: Heterogeneous Environments (Computation Resource \& Data Scale)}

\vspace{3pt}

\noindent\textbf{Problem:} \emph{
%Different medical sites/centers may have significantly different amounts of data and computation resources (e.g., number of GPUs) in real-world practice, so how to reduce its influence on federated training?
This research addresses the disparity in data volumes and computational resources, such as GPU availability, across various medical imaging sites/centers. The focus is on developing strategies to minimize its impact on federated training effectiveness.
}

\vspace{3pt}

In the standard FL algorithm (\ie, FegAvg), each client model conducts a predefined number of training epochs (with equal batches or learning rates) before reaching a synchronization time point when it shares its model with the federation. Different medical sites, however, often have significantly different computation resources and amounts of data (images), which may lead to slow convergence of FL model optimization.
For example, each medical site (client) is supposed to conduct 50 epochs' updates before model uploading. A site with advanced GPUs may take 1 second, while a site with weak computation utility may take 100 seconds. 
Thus, the stronger client will have to spend 99 seconds waiting for weight sharing. 
This will slow down the convergence of the overall federation. 
Aiming at handling this issue in medical imaging analysis tasks, Stripelis~\etal\citep{ISBI_FL1} propose a Semi-Synchronous Training strategy in federated learning and apply it to brain age prediction. 
As shown in Fig.~\ref{fig_unbalance}, each client conducts a variable number of updates (epochs) between synchronization time points which depend on its computational power and data scale. Higher computation power or fewer local data will lead to more local updates (epochs). 

%%%%%%%%%%%%%%%%%%%%%%%%%%%%%%%%%%%%%%%%%%%%%%%%%%%%%%%%%%%%%%
\subsection{Sever-End Learning}\label{Methods-C}
\subsubsection{Sever End: Weight Aggregation}
\vspace{3pt}

\noindent  \textbf{Problem:} \emph{
%How to aggregate the weights of clients properly to avoid performance degradation after each client-server communication?
This research seeks effective strategies for aggregating client weights in a federated learning system, aiming to ensure consistent performance and avoid performance degradation following each client-server communication. 
}

\vspace{3pt}

Weight aggregation in federated learning plays a crucial role by combining the model updates from multiple decentralized clients to form a single, improved global model.
Chen~\etal\citep{chen2022personalized} propose a Progressive Fourier Aggregation strategy at the server end.
Based on previous studies that low-frequency components of parameters
form the basis of deep network capability~\citep{liu2018frequency},
only the low-frequency components are aggregated to share knowledge learned from different clients, while the high-frequency parts are disregarded.
Li~\etal\citep{li2022integrated} consider the training loss of each client as the impact factor of the weight aggregation in FL for COVID-19 detection. The client with relatively bad performance caused by uneven image data will get a smaller weight for the global weight aggregation.

\subsubsection{Sever End: Domain Shift Among Clients}
\vspace{3pt}

\noindent  \textbf{Problem:} \emph{
%The domain shift among clients may cause non-convergence of federated models, so how to avoid this from the server end?
This research addresses the issue of domain shift among clients which can lead to non-convergence of federated models. The focus is on developing server-side solutions that effectively tackle these domain discrepancies, ensuring convergence and stability of the federated models.
}

\vspace{3pt}

The client shift can be handled on the server side during the global model optimization process.
Hosseini~\etal\citep{hosseini2023proportionally} argue that the image heterogeneity between different medical centers (clients) may lead to a biased global model, \ie, a machine learning model that has good performance for some clients while exhibiting inferior performance for other clients.
Thus, they propose a revised optimization objective (motivated by fair resource allocation approaches in wireless network research), to facilitate uniform model performance in histopathology image analysis across all the clients.
In their method, the clients for which the global model has inferior performance will contribute more to the total loss function.
Fan~\etal\citep{fan2021federated} leverage the guided-gradient to optimize the global model. After aggregating all the local weights of the clients, only positive values of the aggregated weights are used to update the global. The authors argue that this is helpful for the global gradient descent to go towards the optimal direction, and the guided-gradient can reflect the most influential regions of the medical images.
Luo~\etal\citep{ISBI_FL4} propose a method called federated learning with shared label distribution (FedSLD) for medical image classification by mitigating label distribution differences among clients.
In their method, it is assumed that the amount of samples of each category (label distribution) is known for the entire federation.
During local model training in each client, a weighted cross-entropy loss is designed as the batch loss.  The weight is computed based on the label distributions in each batch, concerning their label distributions across the entire federation.
%%%%%%%%%%%%%%%%%%%%%%%%%%%%%%%%%%%%%%%%%%%%%%%%%%%%%%%%%%%%%%%%%%%%%%%%%%%%%%%%%%%

%%%%%%%%%%%%%%%%%%%%%%%%%%%%%%%%%%%%%%%% 
\begin{figure}[!tbp]
\setlength{\belowcaptionskip}{0pt}
\setlength{\abovecaptionskip}{0pt}
\setlength{\abovedisplayskip}{-0pt}
\setlength{\belowdisplayskip}{-0pt}
\center
 \includegraphics[width= 0.52\linewidth]{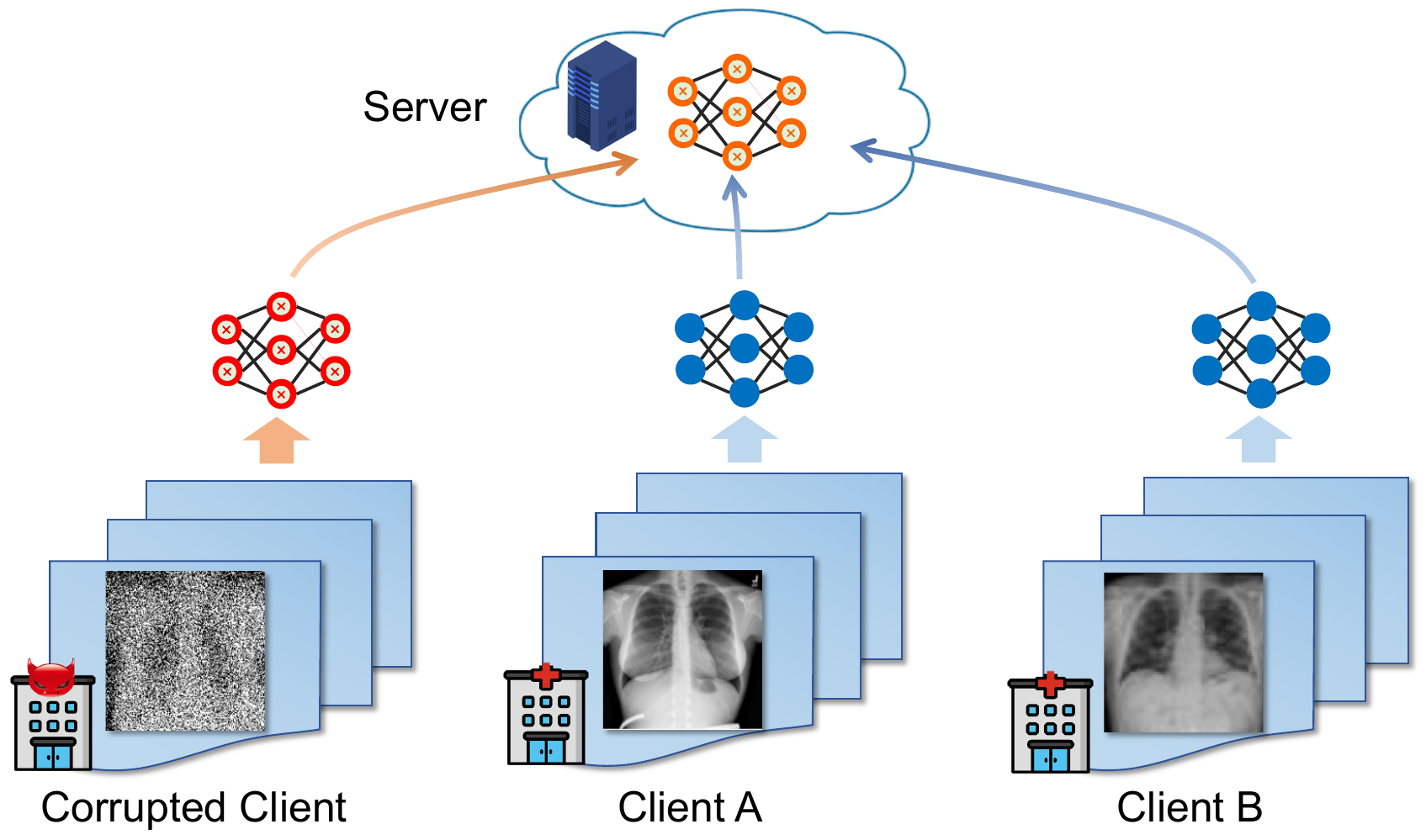}
% \vspace{2}
 \caption{Corrupted clients will lead to a corrupted global model, thus negatively influencing the entire federated learning system.}
 \label{fig_Corrupt}
\end{figure}
%%%%%%%%%%%%%%%%%%%%%%%%%%%
\subsubsection{Sever End: Client Corruption/Anomaly Detection}
\vspace{3pt}

\noindent  \textbf{Problem:} \emph{
%If one or more clients are corrupted by very noisy image labels or malicious attacks, how to avoid its negative influence on the entire federation?
This research investigates how to shield a federated learning system from the impact of clients corrupted by noisy image labels or malicious attacks.
The focus is on developing robust mechanisms that identify and avoid those adverse influences, maintaining the performance of the overall system.
}

\vspace{3pt}

Classic FL framework holds the assumption that all the clients work normally. 
In this context, the term ``normal" means that a client is trained with correctly labeled images or the client is honest without malicious attack.
In real-world practice (as shown in Fig.~\ref{fig_Corrupt}), however, a client may be trained with ``dirty" medical images that have noisy labels, poor scanning qualities, or suffer from poisoning attacks from malicious parties. How to deal with this issue is critical for ensuring the safety of a medical federated learning system. 
Alkhunaizi~\etal\citep{MICCAI-10} propose a sever-end outlier detection method for medical images, called Distance-based Outlier Suppression (DOS), which is robust to client corruption/failure. 
In this method, the weight of each client is calculated based on an anomaly score for the client using Copula-based outlier detection. A client with a high outlier score will get a tiny weight during model aggregation, thus reducing the negative influence of corrupted clients.
Experimental results on clients with noisy labels demonstrate the its effectiveness.

%%%%%%%%%%%%%%%%%%%%%%%%%%%%%%%%%%%%%%%%%%%%%%%%%%%%%%%%%%%%%%%%%%%

\subsection{Client-Server Communication}\label{Methods-D}
\subsubsection{Data Leakage and Attack}
\vspace{3pt}

\noindent  \textbf{Problem:} \emph{
%How to avoid medical image data leakage and privacy violation during the interaction/communication between the server and clients during federated learning?
This research focuses on developing effective methods to prevent medical image data leakage and privacy violations during server-client interactions in federated learning.
}

\vspace{3pt}

Protection of privacy, \ie, ensuring the medical image data of each client are not seen and accessed by other clients/sever, is the main concern of FL systems. 
Prior studies have shown that, even without inter-site data sharing,  pixel-level images can be reconstructed or recovered by the leaked gradients of a machine learning model~\citep{geiping2020inverting,yin2021see,zhu2019deep}. 
Therefore, it is critical to study advanced techniques to proactively avoid data leakage during communication between the server and multiple clients.  
Many studies have focused on this topic in recent years. 

\vspace{3pt}
\noindent \textbf{(1) Partial Weights Sharing.} 
Yang~\etal\citep{yang2021flop} argue that sharing an entire model (network) may not fully protect privacy, and thus propose sharing a partial model for federated learning on medical datasets. 
Specifically, clients only share the feature-learning part of a model for aggregation on the server while keeping the last several layers private.
Similar strategies can also be found in ~\citep{li2019privacy}.

\vspace{3pt}
\noindent \textbf{(2) Differential Privacy.} 
Gradient information of a deep neural work may contain individual privacy that can be reconstructed by malicious parties. 
Differential privacy~\citep{dwork2014algorithmic} could limit the certainty in inferring an individual's presence in the training dataset.
And several recent studies~\citep{li2020multi,lu2022federated,AAAI-1} propose to add Gaussian random noise to the computed gradients on the patients' imaging data in each client/site, thus protecting privacy from the server and other clients.
%%%%%%%%%%%%%%%%%%%%%%%%%%%%%%%%%%%%%%%%%%%%%%%%%%%%%%%%%%%%%%%%%%%%%%%%%%%%%%%%
\vspace{3pt}

\noindent \textbf{(3) Attack and Defense.} 
Kaissis~\etal\citep{PriMIA} apply gradients attack~\citep{geiping2020inverting} to a medical image classification system,
and conduct an empirical study on its capability of reconstructing training images from clients in an FL system.
Hatamizadeh~\etal\citep{hatamizadeh2023gradient} design a gradient inversion algorithm to estimate the running statistics (\ie, mean and variance) of batch normalization layers to match the gradients from real images and the synthesized ones, thus generating synthesized images that are very similar to the original ones.
They further propose a method to measure and visualize
the potential data leakage.

%%%%%%%%%%%%%%%%%%%%%%%%%%%%%%%%%%%%%%%%%%%%%%%%%%%%%%%%%%%%%%%%%%%%%%%%%%%%%%%%%
\subsubsection{Communication Efficiency}
\vspace{3pt}

\noindent  \textbf{Problem:} \emph{
This research is dedicated to formulating strategies that optimize client-server communication, aiming to accelerate the convergence process and ensure more effective model training.
}

\vspace{3pt}

To improve the communication efficiency during FL training, Zhang~\etal\citep{zhang2021dynamic} propose a dynamic fusion-based FL approach for COVID-19 diagnosis.
Their framework dynamically selects the participating clients for weight fusion according to the performance of local client models, and conducts model aggregation based on participating clients' training time.
If a client does not upload its updated model within a certain waiting time, it will be excluded by the central server for this aggregation round.
%%%%%%%%%%%%%%%%%%%%%%%%%%%%%%%%%%%%%%%%%%%%%%%%%%%%%%%%%%%%%%%%%%%%%%%%%%%%%%%%%

%%%%%%%%%%%%%%%%%%%%%%%%%%%%%%%%%%%%%%%%%%%%%%%%%%%%%%%%%%%%%%%%%%%%%%%%%%%%%%%%%%%
\section{Software Platforms and Tools} \label{Software}

In this section, we review several popular and influential federated learning platforms for medical image analysis.
These software platforms provide application interfaces (APIs) for the development of FL systems, which can boost the efficiency and robustness of building large FL systems.

%\vspace{5pt}
%noindent \textbf{PySyft}
\subsection{{PySyft}}
PySyft~\citep{pysyft}\footnote{https://github.com/OpenMined/PySyft} is an open-source FL library enabling secure and private
machine learning by wrapping popular deep learning frameworks.
It is implemented by Python and can run on Linux, MacOS, and Windows systems. 
PySyft has attracted more than 8,000 stars and 1,900 forks on GitHub\footnote{https://github.com}, which shows its popularity.
Budrionis~\etal\citep{budrionis2021benchmarking} carry out an empirical study using PySyft on a medical dataset. %, with experimental results suggest that .  
Their experimental results demonstrate that the performance of machine learning models
trained with federated learning is comparable to those
trained on centralized data.

\subsection{OpenFL}
The Open Federated Learning (OpenFL)\footnote{https://github.com/securefederatedai/openfl} is an open-source FL framework initially developed for use in medical imaging.
The OpenFL is built through a collaboration between Intel and the University of Pennsylvania (UPenn) to develop the Federated Tumor Segmentation (FeTS) platform\footnote{https://www.fets.ai}.
OpenFL supports model training with PyTorch and TensorFlow.
Foley~\etal\citep{foley2022openfl} provide several use cases of OpenFL in medicine, such as tumor segmentation and respiratory distress syndrome prediction.

\subsection{PriMIA}
The Privacy-preserving Medical Image Analysis (PriMIA)~\citep{PriMIA} is an open-source framework for privacy-preserving decentralized deep learning with medical images.
PriMIA is built upon the PySyft ecosystem which supports Python and PyTorch for deep learning development.
It is compatible with a wide range of medical imaging data formats.
The source code, documentation as well as publicly available data can be found online (\url{https://zenodo.org/record/4545599}).
For example, Kaissis~\etal\citep{PriMIA} use PriMIA to perform classification on pediatric chest X-rays and achieve good results.

\subsection{Fed-BioMed}
Fed-BioMed\footnote{https://fedbiomed.gitlabpages.inria.fr} is an open-source federated learning software for real-world medical applications.
It is developed by Python and supports multiple machine learning toolkits such as PyTorch, Scikit-Learn, and NumPy.
It can also be used in cooperation with PySyft. 
Silva~\etal\citep{silva2020fed} use Fed-BioMed to conduct multi-center analysis for structural brain imaging data (MRI) across different datasets and verify its effectiveness.

Due to the increasing and extensive influence of federated learning, many software platforms and frameworks have been proposed to date. More comparative reviews and evaluations can be found in~\citep{li2021survey,kholod2020open}.

%%%%%%%%%%%%%%%%%%%%%%%%%%%%%%%%%%%%%%%%%%%%%%%%%%%%%%%%%%%%%%%%%%%%%%%%%%%%%%%%%%%

%%%%%%%%%%%%%%%%%%%%%%%%%%%%%%%%%%%%%%%%%%%%%%%%%%%%%%%%%%%%%%%%%%%%%%%%%%%%%%%
\section{Medical Image Datasets for Federated Learning} \label{Datasets}
In this section, we introduce the benchmark datasets that have been commonly used in federated learning for medical image analysis.
For clarity, these datasets are presented in terms of different research objects/organs.

\subsection{Medical Image Data Usage Overview}
For most existing FL research in medical image analysis, there are typically two ways of using different imaging datasets for simulation and experiment.
The first way is to directly use databases from different medical sites/centers~\citep{li2020multi,dayan2021federated}. These databases are typically research projects that are built through multi-center cooperation.
Thus, they are ideal choices to set up a FL simulation environment. 
Another popular way to build an FL experiment platform is to split a very large-scale medical image dataset into several subsets~\citep{ISBI_FL2,MICCAI-10}, where each subset is treated as a client dataset.

\subsection{Brain Images}
\subsubsection{ADNI}
The Alzheimer's Disease Neuroimaging Initiative
(ADNI)~\citep{ADNI,ADNI2} is the largest and most influential benchmark for the research of
Alzheimer's Disease (AD), %
including ADNI-1, ADNI-2, ADNI-GO and ADNI-3. 
Structural brain MRI, functional MRI, and positron emission tomography (PET) from 1,900+ subjects and 59 centers are provided for analysis and research.

\subsubsection{ABIDE}
Autism Brain Imaging Data Exchange (ABIDE) initiative~\citep{ABIDE} is a benchmark database for research on Autism spectrum disorder.
ABIDE contains both structural and functional brain images independently collected from more than 24 imaging laboratories/sites around the world.

\subsubsection{BraTS}
Multimodal Brain Tumor Image Segmentation Benchmark (BraTS)~\citep{braTS} is a benchmark dataset for brain tumor segmentation.
BraTS is updated regularly for the Brain Tumor Segmentation Challenge\footnote{https://www.med.upenn.edu/cbica/brats}.
It contains brain MRIs acquired by various scanners from around 19 independent institutions.

\subsubsection{RSNA Brain CT}
Radiological Society of North America (RSNA)~\citep{RSNA} is a large-scale multi-institutional CT dataset for intracranial hemorrhage detection.
RSNA contains 874,035 images which are compiled and archived from three different institutions, \ie,
Stanford University (Palo Alto, USA), Thomas Jefferson University
Hospital (Philadelphia, USA), and Universidade Federal
de São Paulo (São Paulo, Brazil).

\subsubsection{UK Biobank}

UK Biobank~\citep{UK-Biobank} is a large-scale, influential biomedical database and research resource containing genetic and health data from half a million participants.
As for imaging data, it has four imaging centers and contains valuable brain scans and cardiac MRI information.
As a large-scale database with multiple imaging centers, UK Biobank can contribute to varied research areas in medical image analysis, such as federated learning and domain adaptation.

\subsubsection{IXI}
IXI Dataset\footnote{https://brain-development.org/ixi-dataset} consists of around 600 MR images from healthy subjects.
All the images are acquired from three different hospitals (using different scanners or scanning parameters) in London.

%%%%%%%%%%%%%%%%%%%%%%%%%%%%%%%%%%%%%%%%%%%%%%%%%%%%%%%%%%%%%%%%%%%%%%%
\subsection{Chest/Lung/Heart Images}
\subsubsection{CheXpert}
CheXpert~\citep{chexpert} is a large-scale dataset including 224,316 chest radiographs of 65,240 patients. These images are acquired from Stanford University Medical Center.

\subsubsection{ChestX-ray}
The ChestX-ray (also known as ChestX-ray14)\footnote{https://www.kaggle.com/datasets/nih-chest-xrays/data} is a large and publicly-available medical image dataset that contains 112,120 X-ray images (in frontal-view) of 30,805 patients with 14 disease labels. It is expanded from the ChestX-ray8 dataset~\citep{chestxray14} by adding six thorax diseases, including Edema, Emphysema, Fibrosis, Hernia, Pleural, and Thickening.

\subsubsection{COVID-19 Chest X-ray}
The COVID-19 Chest X-ray (also known as COVID-19 CXR)~\citep{COVID-19-CXR}\footnote{https://www.kaggle.com/datasets/tawsifurrahman/covid19-radiography-database} is a publicly-available database of chest X-ray images, containing 3,616 COVID-19 positive cases, 10,192 normal controls, 6,012 lung opacity (non-COVID infection), and 1,345 viral pneumonia cases.

\subsubsection{COVIDx}
The COVIDx dataset~\citep{covidx} is a large-scale and fully accessible database comprising 13,975 chest X-ray images of 13,870 patients.
COVIDx includes 358 chest X-ray images from 266 COVID-19 patient cases, 8,066 normal cases, and 5,538 non-COVID-19 pneumonia cases.

\subsubsection{ACDC}
Automatic Cardiac Diagnosis Challenge (ACDC)~\citep{ACDC} is a large publicly available and fully annotated dataset for cardiac MRI assessment.
This dataset consists of 150 patients who are divided into 5 categories in terms of well-defined characteristics based on physiological parameters.

\subsubsection{M$\&$M}
Multi-Center, Multi-Vendor, and Multi-Disease Cardiac Segmentation (M$\&$Ms) Challenge~\citep{MM}\footnote{https://www.ub.edu/mnms} is a publicly available cardiac MRI dataset.
This dataset contains 375 participants from 6 different hospitals in Spain, Canada, and Germany. All the cardiac MRIs are acquired by 4 different scanners (\ie, GE, Siemens, Philips, and Canon).

%%%%%%%%%%%%%%%%%%%%%%%%%%%%%%%%%%%%%%%%%%%%%%%%%%%%%%%%%%%%%%%%%%%%%
\subsection{Skin Images}
\subsubsection{HAM10000}
The ``Human Against Machine with 10000 training images" (HAM10000)~\citep{ham10000}\footnote{https://www.kaggle.com/datasets/kmader/skin-cancer-mnist-ham10000} is a popular large-scale dataset for diagnosis of pigmented skin lesions.
It consists of 10,015 dermatoscopic images from different sources.
Cases in this dataset include a collection of all representative diagnostic categories of pigmented lesions.

\subsubsection{ISIC}
The International Skin Imaging Collaboration (ISIC) challenge dataset~\citep{ISIC}\footnote{https://challenge.isic-archive.com/data} is a large-scale database, containing a series of challenges for skin lesion image analysis.
ISIC has become a standard benchmark dataset for dermatoscopic image analysis.

\subsection{Others} 

\subsubsection{Eye: Kaggle Diabetic Retinopathy (Retina)}
The Kaggle Diabetic Retinopathy (Retina)\footnote{https://www.kaggle.com/competitions/diabetic-retinopathy-detection/data} is a large-scale dataset of color digital retinal fundus images for diabetic retinopathy detection.
It includes 17,563 pairs of color digital retinal fundus images. 
Each image in this dataset is provided a label (a rated scale from 0 to 4) in terms of the presence of diabetic retinopathy, where 0 to 4 represents no, mild, moderate, severe, and proliferative diabetic retinopathy, respectively. 

\subsubsection{Abdomen: PROMISE12}
The MICCAI 2012 Prostate MR Image Segmentation challenge dataset (PROMISE12)\citep{PROMISE12} is a publicly
available dataset for the evaluation of prostate MRI segmentation methods. 
It consists of 100 prostate MRIs acquired by different scanners from 4 independent medical centers,
including University College London in the United Kingdom, Haukeland University Hospital in Norway, the Radboud University Nijmegen Medical Centre in the Netherlands, and the Beth Israel Deaconess Medical Center in the USA.

\subsubsection{Histology: TCGA}
The Cancer Genome Atlas (TCGA)~\citep{TCGA}\footnote{https://www.cancer.gov/ccg/research/genome-sequencing/tcga} is a large-scale landmark cancer genomics database.
Whole-slide images for normal controls and cancers are provided for histology and microscopy research. 

\subsubsection{Knee: fastMRI}
The fastMRI~\citep{fastMRI,fastMRI2}\footnote{https://fastmri.med.nyu.edu} is a large-scale dataset for medical image reconstruction using machine learning approaches.
This dataset contains more than 1,500 knee MRIs (1.5 and 3 Tesla) and DICOM images from 10,000 clinical knee MRIs (1.5 and 3 Tesla).

\subsubsection{MedMNIST}
MedMNIST~\citep{MedMNIST} is a dataset for medical image classification.
Similar to the MNIST dataset\footnote{http://yann.lecun.com/exdb/mnist}, all the images in the MedMNIST are stored as the size of 28 $\times$ 28.
The MedMNIST includes 10 pre-processed subsets, covering primary modalities (\eg, MR, CT, X-ray, Ultrasound, OCT).
As a lightweight dataset with diversity, MedMNIST is good for rapid prototyping machine learning algorithms. 
%%%%%%%%%%%%%%%%%%%%%%%%%%%%%%%%%%%%%%%%%%%%%%%%%%%%%%%%%%%%%%%%%%%%%%
%%%%%%%%%%%%%%%%%%%%%%%%%%%%%%%%%%%%%%%%%%%%%%%%%%%%%%%%%%%%%%%%%%%%%%%%%%%%%%%%%%%%
%\begin{center}
{\scriptsize
 
\begin{longtable}{p{4.5cm} p{4.5cm} p{3.0cm} p{2.0cm} p{2.5cm}}%[!tbp]
%\footnotesize
%\scriptsize
\caption{Benchmark datasets for federated learning in medical image analysis.}
\label{tab:dataset}\\
%\centering
%\setlength{\tabcolsep}{0.5mm}{
%\begin{tabular}{p{3.6cm} p{4.5cm} p{3.0cm} p{2.5cm} p{2.5cm}}%{lllll} 
\hline%\toprule
Reference  &Task   &Dataset   &Modality   &Learning Model\\
\hline%\midrule
\endfirsthead

\hline%\toprule
Reference  &Task   &Dataset   &Modality   &Model\\
\hline%\midrule
\endhead

\textbf{Brain}          &~         &~         &~        &~     \\
Peng~\etal (2022)\citep{peng2022fedni}
&ASD, AD classification     &ABIDE~\citep{ABIDE}, ADNI~\citep{ADNI} 
&fMRI                       &GCN      \\

G{\"u}rler~\etal (2022)\citep{gurler2022federated}  
&Brain connectivity prediction     &OASIS~\citep{OASIS} 
&MRI                    &GNN      \\

Islam~\etal (2022)\citep{islam2022effectiveness}  
&Brain tumor classification     &UK Data Service~\citep{UK_data_service} 
&MRI                    &CNN      \\

Dinsdale~\etal (2022)\citep{MICCAI-12}  
&Age prediction     &ABIDE~\citep{ABIDE} 
&MRI                   &CNN (VGG)       \\

Jiang~\etal (2022)\citep{MICCAI-8}  
&Intracranial hemorrhage diagnosis     &RSNA~\citep{RSNA} 
&CT                                    &CNN (DenseNet)       \\

Stripelis~\etal(2021)\citep{ISBI_FL1}  
&Brain age prediction     &UK Biobank~\citep{UK-Biobank}     
&MRI                                &CNN   \\

Liu~\etal(2021)\citep{MICCAI-4}  
&Intracranial hemorrhage diagnosis     &RSNA~\citep{RSNA} 
&CT                                    &CNN (DenseNet)       \\

Fan~\etal (2021)\citep{fan2021federated}  
&ASD classification     &ABIDE~\citep{ABIDE} 
&MRI                    &CNN      \\

Li~\etal (2020)\citep{li2020multi}  
&ASD classification     &ABIDE~\citep{ABIDE} 
&fMRI                   &MLP      \\

Sheller~\etal (2019)\citep{sheller2019multi}
&Brain tumor segmentation      &BraTS~\citep{braTS} 
&MRI                           &U-Net      \\

Li~\etal (2019)\citep{li2019privacy}
&Brain tumor segmentation      &BraTS~\citep{braTS} 
&MRI                           &CNN      \\

\hline
\textbf{Chest}          	&~     &~         &~    &~       \\
Hatamizadeh~\etal(2023)\citep{hatamizadeh2023gradient}
&Image generation (attack)           
&COVID CXR~\citep{COVID-19-CXR} \newline ChestX-ray14~\citep{chestxray14}   
&Chest X-ray               &CNN (ResNet)     \\

Yan~\etal(2023)\citep{yan2023label}
&Classification            &COVID-FL~\citep{yan2023label}    
&Chest X-ray               &Transformer     \\

Alkhunaizi~\etal (2022)\citep{MICCAI-10}  
&Classification        &CheXpert~\citep{chexpert} 
&Chest X-ray           &CNN       \\

Dong~\etal (2022)\citep{MICCAI-14}  
&Classification        &ChestX-ray14~\citep{chestxray14} 
&Chest X-ray           &CNN       \\

Chakravarty~\etal(2021)\citep{ISBI_FL2} 
&Classification            &CheXpert~\citep{chexpert}     
&Chest X-ray               &CNN, GNN     \\

\textbf{Lung}          	&~     &~         &~     &~      \\
Yang~\etal (2021)\citep{yang2021flop} 
&COVID-19 diagnosis    &COVIDx \citep{covidx}       
&Chest X-ray           &CNN   \\

Feki~\etal (2021)\citep{feki2021federated} 
&COVID-19 diagnosis    &Local dataset%\footnote{https://github.com/ieee8023/covid-chestxray-dataset}       
&Chest X-ray           &CNN   \\

Kumar~\etal (2021)\citep{kumar2021medisecfed} 
&COVID-19 diagnosis    &COVID-19 CXR\citep{COVID-19-CXR}       
&Chest X-ray           &CNN   \\

Dong~\etal (2021)\citep{MICCAI-2} 
&COVID-19 diagnosis    &COVID-19 CXR\citep{COVID-19-CXR}       
&Chest X-ray           &CNN   \\

Yang~\etal (2021)\citep{yang2021federated} 
&Segmentation    &Local dataset      
&CT                    &CNN   \\

\textbf{Heart}          	&~     &~         &~    &~       \\
Linardos~\etal (2022)\citep{linardos2022federated}  
&Cardiac diagnosis      &ACDC\citep{ACDC}, M\&M\citep{MM}       
&MRI                    &CNN       \\

Qi~\etal (2022)\citep{MICCAI-8}  
&Cardiac segmentation      &M\&M\citep{MM}, Emidec\citep{Emidec}       
&MRI                       &U-Net       \\

Li~\etal (2021)\citep{MICCAI-1}   
&Cardiac image synthesis         &Local dataset       
&CT                              &GAN     \\

Wu~\etal (2021)\citep{wu2022distributed}  
&Cardiac segmentation      &ACDC\citep{ACDC}       
&MRI                       &U-Net       \\

\hline
\textbf{Breast}       	&~       &~       &~       &~     \\
Agbley~\etal (2023)\citep{agbley2023federated}  
&Breast tumor classification    &BreakHis~\citep{breakhis}    
&Pathology      &CNN   \\

Wicaksana~\etal (2022)\citep{fedmix}  
&Breast tumor segmentation     &BUS~\citep{BUS}, BUSIS~\citep{BUSIS}, UDIAT~\citep{UDIAT}   
&Ultrasound    &U-Net   \\

\hline
\textbf{Skin}         	&~        &~          &~     &~         \\
Yan~\etal (2023)\citep{yan2023label}  
&Skin lesion classification     &ISIC~\citep{ISIC}%~\footnote{https://challenge.isic-archive.com/data/}
&Dermoscopy                     &Transformer       \\

Wicaksana~\etal (2022)\citep{wicaksana2022customized}  
&Skin lesion classification     &HAM10000~\citep{ham10000}
&Dermoscopy                     &CNN       \\

Alkhunaizi~\etal (2022)\citep{MICCAI-10}  
&Skin lesion classification     &HAM10000~\citep{ham10000}
&Dermoscopy                     &CNN       \\

Jiang~\etal (2022)\citep{MICCAI-8}  
&Skin lesion classification     &HAM10000~\citep{ham10000} 
&Dermoscopy                             &CNN (DenseNet) \\

Liu~\etal (2021)\citep{MICCAI-4}  
&Skin lesion classification     &HAM10000~\citep{ham10000} 
&Dermoscopy                     &CNN (DenseNet)       \\

Bdair~\etal (2021)\citep{MICCAI-6}  
&Skin lesion classification     &HAM10000~\citep{ham10000} \newline ISIC19~\citep{ISIC19} \newline Derm7pt~\citep{Derm7pt} \newline PAD-UFES~\citep{PAD-UFES}
&Dermoscopy                     &CNN (EfficientNet)       \\

Chen~\etal (2021)\citep{MICCAI-7}  
&Skin lesion classification     &HAM10000~\citep{ham10000} \newline ISIC17~\citep{ISIC17} 
&Dermoscopy                     &CNN (VGG)       \\

\hline
\textbf{Eye}         		&~        &~         &~      &~      \\
Yan~\etal (2023)\citep{yan2023label}
&Diabetic classification        &Retina\citep{Retina}       
&Color retinal image            &Transformer  \\

Qiu~\etal (2023)\citep{qiu2023federated}
&Fundus segmentation        & RIM-ONE~\citep{RIM-ONE}    
&Color retinal image        &CNN (MobileNet)  \\

Wang~\etal (2023)\citep{wang2023feddp}
&Fundus segmentation        & RIF~\citep{liu2021feddg}   
&Color retinal image        & Transformer \\

Qu~\etal (2022)\citep{qu2022handling}   
&Diabetic classification        &Retina~\citep{Retina}       
&Color retinal image            &VAE, CNN   \\

\hline
%%%%%%%%%%%%%%%%%%%%%%%%%%%%%%%%%%%%%%%%%
\textbf{Abdomen}   &~        &~         &~     &~     \\
Zhu~\etal(2023)\citep{zhu2023feddm}
&Prostate segmentation       &PROMISE12~\citep{PROMISE12} \newline NCI-ISBI 2013~\citep{NCI-ISBI}  
&MRI                         &U-Net   \\

Qiu~\etal (2023)\citep{qiu2023federated}
&Prostate segmentation        &PROMISE12~\citep{PROMISE12}     
&MRI        &CNN (MobileNet)  \\

Xu~\etal(2023)\citep{xu2023federated}
&Tumor segmentation      &LiTS~\citep{LiTS}
&CT                      &U-Net   \\

Wicaksana~\etal (2022)\citep{wicaksana2022customized}  
&Cancer classification     &ProstateX~\citep{ProstateX} 
&MRI         &CNN       \\

Luo~\etal(2022)\citep{ISBI_FL4}
&Cancer classification       &OrganMNIST~\citep{MedMNIST}   
&CT     &CNN   \\

Liu~\etal(2022)\citep{MICCAI-15}
&Polyp detection       &GLRC~\citep{GLRC}   
&Colonoscopy           &CNN   \\

Yan~\etal(2021)\citep{yan2020variation}
&Cancer classification       &ProstateX~\citep{ProstateX}  
&MRI    &GAN   \\

Roth~\etal(2021)\citep{MICCAI-5}
&Prostate segmentation       &MSD-Prostate~\citep{MSD-Prostate} \newline PROMISE12~\citep{PROMISE12} \newline ProstateX~\citep{ProstateX} \newline NCI-ISBI 2013~\citep{NCI-ISBI}  
&MRI                         &U-Net   \\
\hline
%%%%%%%%%%%%%%%%%%%%%%%%%%%%%%%%%%%%%%%%%
\textbf{Histology}         &~        &~         &~    &~       \\
Hosseini~\etal(2023)\citep{hosseini2023proportionally}
&Cancer classification       &TCGA~\citep{TCGA}
&Pathology                &CNN (DenseNet)   \\

du Terrail~\etal(2023)\citep{du2023federated}
&Cancer classification       &Local dataset
&Pathology                   &CNN   \\

Lu~\etal(2022)\citep{lu2022federated}  
&Cancer classification       &TCGA~\citep{TCGA}
&Pathology    &CNN   \\

Adnan~\etal(2022)\citep{Exp1}  
&Cancer classification       &TCGA~\citep{TCGA}
&Pathology    &CNN (DenseNet)   \\

Luo~\etal(2022)\citep{ISBI_FL4}
&Cancer classification       &PathMNIST~\citep{MedMNIST}    
&Pathology     &CNN    \\

Wagner~\etal(2022)\citep{MICCAI-13}
&Image harmonization      &PESO~\citep{PESO}    
&Pathology     &GAN    \\

Ke~\etal(2021)\citep{ISBI_FL3}  
&Image harmonization       &TCGA~\citep{TCGA}      
&Pathology    &GAN    \\

\hline
\textbf{Others}   &~        &~         &~    &~      \\
Feng~\etal (2022)\citep{feng2022specificity}
&MRI reconstruction          &fastMRI~\citep{fastMRI}\newline BraTS~\citep{braTS}
&MRI                        &U-Net     \\

Elmas~\etal(2022)\citep{elmas2022federated}   &MRI reconstruction      
&fastMRI, BraTS, IXI%~\citep{IXI}  
&MRI    &GAN    \\

Guo~\etal(2021)\citep{guo2021multi}   &MRI reconstruction       &fastMRI, BraTS, IXI  &MRI    &U-Net    \\

\bottomrule
%\end{tabular}
%}

\end{longtable}
}
%\end{center}
%%%%%%%%%%%%%%%%%%%%%%%%%%%%%%%%%%%%%%%%%%%%%%%%%%%%%%%%%%%%%%%%%%%%%%%%%

\section{Experiment} \label{Experiment}
\if false
To empirically evaluate the federated learning performance of different approaches for medical image analysis, we conduct an experiment to assess several representative FL methods and some methods with diverse settings on a popular benchmark dataset.

\subsection{Dataset}
We conduct the experiment on the popular benchmark ADNI dataset~\citep{ADNI,ADNI2}. 
Two studies/phases in ADNI (\ie, ADNI-1 and ADNI-2) with baseline data are used as two client datasets, where subjects that appear in both ADNI-1 and ADNI-2 are removed from ADNI-2 for independent evaluation. 
Specifically, ADNI-1 consists of 1.5T T1-weighted structural MRIs of 428 subjects (including 199 patients with AD and 229 normal controls (NCs)), while ADNI-2 contains 3.0T T1-weighted structural MRIs of 360 subjects (including 159 AD patients and 201 NC subjects).
We use brain regions-of-interest (ROIs) as the features to represent each MRI.
The ROI features are calculated based on the mean gray matter volumes of 90 brain regions defined in the AAL atlas~\citep{AAL}. 
In all experiments, for each client, 80\% of the dataset is randomly selected to construct the training set, while the remaining 20\% samples are used for test. 
To avoid bias caused by random partition, the random partition process is repeated five times, and we record and report the mean and standard deviation results. 
\fi

To evaluate federated learning (FL) performance in medical image analysis, we compared various FL approaches using the ADNI dataset~\citep{ADNI,ADNI2}. This dataset comprises two subsets: ADNI-1, featuring 1.5T T1-weighted MRIs from 428 subjects (199 with Alzheimer's Disease (AD) and 229 normal controls (NCs)), and ADNI-2, with 3.0T T1-weighted MRIs from 360 subjects (159 AD patients and 201 NCs). Duplicate subjects across ADNI-1 and ADNI-2 were excluded for independence. Analysis focused on 90 brain regions-of-interest (ROIs) based on the AAL atlas~\citep{AAL}, using mean gray matter volumes as features. For each experiment, 80\% of the data was used for training and 20\% for testing, with this split performed five times to ensure reliability. Results include mean and standard deviation values to account for variability.
%%%%%%%%%%%%%%%%%%%%%%%%%%%%%%%%%%%%%%%%%%%%%%%%%%%%%%%%%%%%%%%%%

\subsection{Experimental Setup}
%The task here is AD vs. NC classification based on structural MRI data. 
The task here is AD vs. NC classification.
We use four metrics for performance evaluation, including classification accuracy (ACC), sensitivity (SEN), specificity (SPE), and area under the ROC curve (AUC). 
%Logistic Regression (with model weight $\mathbf{w}$) 
Logistic Regression is used as the machine learning model for each FL setting, which has been widely used in medical imaging analysis~\citep{logistic1,logistic2,logistic3,logistic4}.
%\subsection{Federated Learning Settings for Comparison}
We compare 3 conventional machine learning and 3  popular FL methods in our study, with details given below. 

\begin{figure}[t]%[!tbp]
\setlength{\belowcaptionskip}{-2pt}
\setlength{\abovecaptionskip}{0pt}
\setlength{\abovedisplayskip}{0pt}
\setlength{\belowdisplayskip}{0pt}
\center
 \includegraphics[width= 0.5\linewidth]{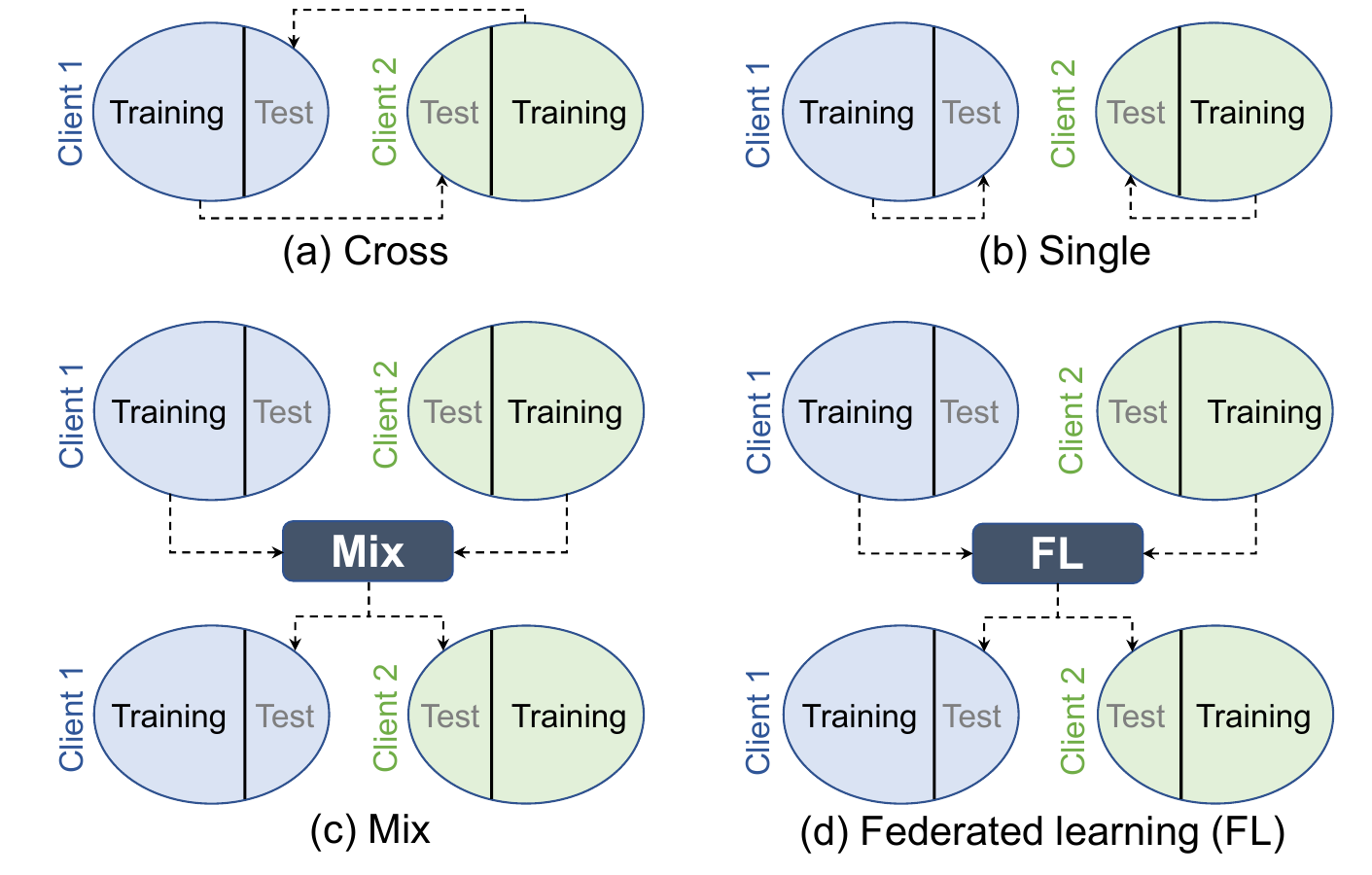}
% \vspace{2}
 \caption{Different settings for performance comparison.}
 \label{fig_Settings}
\end{figure}

(1) \textbf{Cross}.
Training is conducted on one client dataset and then the trained model is directly tested on the data of the other client, as shown in Fig.~\ref{fig_Settings}~(a).
Specifically, ADNI-1 is used as the training set (denoted as ADNI1-tr), then the trained model is tested on ADNI-2. 
ADNI-2 is used as the training set (denoted as ADNI2-tr), then the trained model is evaluated on ADNI-1.

(2) \textbf{Single}.
Training and testing are conducted within each client dataset separately, as shown in Fig.~\ref{fig_Settings}~(b). 
In each client, 80\% of the data is used for training while the other is used for testing.

(3) \textbf{Mix}.
All the training data in each client are pooled together for training a model, then the trained model is evaluated on the test data of all the clients, as shown in Fig.~\ref{fig_Settings}~(c).
Note this strategy needs to share data, and thus, could not preserve privacy.

(4) \textbf{FedAVG}~\citep{FL,FedAvg2}.
Each client trains its own model, then their model weights (\eg, the weight $\mathbf{w}$ of logistic regression) are aggregated to calculate a global model.
The final trained global model is tested on all the test data in each client, as shown in Fig.~\ref{fig_Settings}~(d). 
The number of iterations for local model training is set to 10.

(5) \textbf{FedSGD}~\citep{FL}.
Each client trains a local model, and then the gradients from each client are aggregated to calculate a global model.
The global model is then applied to all the test data in each client for assessment, as shown in Fig.~\ref{fig_Settings}~(d).
The number of iterations for local model training is set to 10.

(6) \textbf{FedProx}~\citep{FedProx}. 
Every client trains its own model with an additional proximal term (the coefficient $\mu$ is set to 0.1).
Local training is conducted only once.
The model weights of each client are aggregated to get a global model.
The trained global model is then assessed on the test data in each client, as shown in Fig.~\ref{fig_Settings}~(d).

%%%%%%%%%%%%%%%%%%%%%%%%%%%%%%%%%%%%%%%%%%%%%%%%%%%%%%%%%%%%%%%

%%%%%%%%%%%%%%%%%%%%%%%%%%%%%%%%%%%%%%%% 

%%%%%%%%%%%%%%%%%%%%%%%%%%%%%%%%%%%%%%%%%%%%%%%%%%%%%%%%%%%%%%%
%%%%%%%%%%%%%%%%%%%%%%%%%%%%%%%%%%%%%%%%%%%%%%%%%%%%%%%%%%%%%%%%%%%%%%%%%%%%%%%%
\begin{table*}[t]%[!tbp]
\setlength{\belowcaptionskip}{0pt}
\setlength\abovedisplayskip{0pt}
\setlength\belowdisplayskip{0pt}
\renewcommand\arraystretch{0.8}
%\tiny
\scriptsize
\setlength{\tabcolsep}{0.5pt}
\begin{center}
\caption{Classification results (mean$\pm$standard deviation) of different federated learning settings in terms of four metrics. ADNI1-tr: ADNI-1 is adopted as the training set. ADNI2-tr: ADNI-2 is used as the training set.
}
%\vspace{5pt}
\begin{tabularx}{\textwidth}{ X X X X X X }
%\begin{tabularx}{1\hsize}{@{}@{\extracolsep{\fill}}l| l| c c c c @{}}
\toprule %[1.2pt]
Client    &Method   &ACC   &SEN        &SPE        &AUC \\
\midrule

\multirow{7}*{ADNI-1 }  
  &ADNI1-tr &--          &--      &--      &--  \\
~ &ADNI2-tr &0.818            &0.809      &0.825      &0.886 \\
~ &Single   &0.844$\pm$0.013  &0.786$\pm$0.045  &0.895$\pm$0.040  &0.889$\pm$0.018 \\
~ &Mix      &0.870$\pm$0.017  &0.823$\pm$0.045  &0.923$\pm$0.050  &0.901$\pm$0.018   \\
~ &FedAvg   &0.860$\pm$0.028  &0.783$\pm$0.025  &0.901$\pm$0.049  &0.897$\pm$0.013  \\
~ &FedSGD   &0.823$\pm$0.030  &0.752$\pm$0.047  &0.882$\pm$0.011  &0.880$\pm$0.034 \\
~ &FedProx  &0.858$\pm$0.031  &0.815$\pm$0.077  &0.896$\pm$0.034  &0.900$\pm$0.044  \\

\midrule

\multirow{7}*{ADNI-2} 
  &ADNI1-tr  &0.811     &0.623      &0.960      &0.885  \\
~ &ADNI2-tr  &--     &--      &--      &-- \\
~ &Single    &0.828$\pm$0.016  &0.750$\pm$0.045  &0.890$\pm$0.046  &0.863$\pm$0.043 \\
~ &Mix       &0.872$\pm$0.012  &0.843$\pm$0.048  &0.898$\pm$0.038  &0.910$\pm$0.013 \\
~ &FedAvg    &0.842$\pm$0.021  &0.823$\pm$0.018  &0.864$\pm$0.038  &0.907$\pm$0.017  \\
~ &FedSGD    &0.844$\pm$0.039  &0.819$\pm$0.066  &0.871$\pm$0.056  &0.908$\pm$0.040  \\
~ &FedProx   &0.856$\pm$0.045  &0.845$\pm$0.072  &0.861$\pm$0.047  &0.908$\pm$0.036  \\

\bottomrule %[1.2pt]
\end{tabularx}
\label{AD_task}
\end{center}
\end{table*}
%%%%%%%%%%%%%%%%%%%%%%%%%%%%%%%%%%%%%%%%%%%%%%%%%%%%%%%%%%

%%%%%%%%%%%%%%%%%%%%%%%%%%%%%%%%%%%%%%%%%%%%%%%%%%%%%%%%%%%%%%%%%%
\subsection{Result and Analysis}
The classification result of different methods is shown in Table~\ref{AD_task}.
In the ``Cross" setting, the client dataset for training is denoted as ``$<$client$>$ (tr)". 
Since there is only one test dataset in this setting, no standard deviation is reported.

From Table~\ref{AD_task}, we can get the following observations.
1) The ``mix" strategy has the best performance.
This is because it combines all the training data of the clients and the learning model can get access to the largest amount of data information than the other methods.
2) The ``cross" strategy has the worst performance. 
This should be caused by the well-known ``domain shift" problem. Since ADNI-1 and ADNI-2 have different scanning parameters, then directly transferring a model may not achieve good classification results. 
3) Federated learning methods achieve satisfactory performance.
This can be explained by FL can leverage more data information than the baseline methods (\ie, ``cross" and ``single"), even without cross-site data sharing.
4) Among the FL methods, we find that aggregation of model weights (\ie, FedAvg, FedProx) can be more advantageous than a fusion of the gradients of each client model (\ie, FedSGD).

\if false
In this experiment, we conduct an empirical evaluation of FL algorithms on a benchmark dataset.
The result can provide some valuable insights for medical imaging researchers.
It should be acknowledged that the results may not generalize to all medical imaging scenarios, as different datasets and clinical settings could yield varying outcomes.
\fi
%%%%%%%%%%%%%%%%%%%%%%%%%%%%%%%%%%%%%%%%%%%%%%%%%%%%%%%%%%%%%%%%%%%%%%%
%%%%%%%%%%%%%%%%%%%%%%%%%%%%%%%%%%%%%%%%%%%%%%%%%%%%%%%%%%%%%%%%%%%%%%%%%

%%%%%%%%%%%%%%%%%%%%%%%%%%%%%%%%%%%%%%%%%%%%%%%%%%%%%%%%%%%%%%%%%%%%%%%%%
\section{Discussion} \label{Discussions}
\subsection{Challenges of FL for Medical Image Analysis}

\subsubsection{Data Heterogeneity Among Clients}
Data heterogeneity is widespread in real-world medical image sites.
Such heterogeneity can hardly be avoided in practice due to the following factors. 
1) Medical images from different sites/datasets are typically acquired by different scanners or scanning protocols. %even under different modalities.
2) Patients in different sites/hospitals have different distributions.
The heterogeneous data distribution, \ie, ``domain shift" or ``client shift", may cause significant degradation or biased performance of a federated learning system.
How to alleviate the negative influence of data heterogeneity is one of the most important and challenging research problems for federated learning in medical imaging.

%%%%%%%%%%%%%%%%%%%%%%%%%%%%%%%%%%%%%%%%%%%%%%%%%%%%%%%%%%%%%%%%%%%%%%%
\subsubsection{Privacy Leakage/Poisoning Attacks}
In classic FL, only the model parameters are exchanged and updated without data sharing.
This is considered an effective way of privacy protection.
But further research reveals that FL still faces privacy and security risks, including privacy leakage~\citep{geiping2020inverting,yin2021see,zhu2019deep} and poisoning attacks~\citep{lyu2022privacy,xia2023poisoning}.
These issues can happen at both the server end and the client end.
Since an FL system contains the communication and interaction of many entities/parties, how to effectively protect individual privacy and data %and defend privacy and 
security is a very challenging problem.
%%%%%%%%%%%%%%%%%%%%%%%%%%%%%%%%%%%%%%%%%%%%%%%%%%%%%%%%%%%%%%%
%%%%%%%%%%%%%%%%%%%%%%%%%%%%%%%%%%%%%%%%%%%%%%%%%%%%%%%%%%%%%%%%%
\subsubsection{Technological Limitations}
While a majority of research is centered on algorithm design for various medical applications, the practical implementation of FL systems encounters significant technological hurdles. For instance, certain FL algorithms demand substantial computational resources, posing challenges for the underlying hardware infrastructure. Furthermore, addressing communication costs, optimizing network resource allocation, and ensuring synchronization are all formidable obstacles when striving to construct a robust and functional FL system in real life.

\subsubsection{Long-Term Viability of FL-based Medical Image Analysis}
Federated Learning is not just a novel machine learning algorithm; it represents a dynamic and systematic approach to engineering.
It is imperative to focus on the long-term viability of FL systems when applied to medical image analysis. This involves addressing critical issues such as scalability, sustainability, and evolving regulations. In real-world scenarios, unforeseen challenges emerge, such as clients leaving or joining the training process, as well as unexpected technical and connectivity issues. Effectively managing an FL system for robust and stable long-term medical image analysis is a complex endeavor.
%%%%%%%%%%%%%%%%%%%%%%%%%%%%%%%%%%%%%%%%%%%%%%%%%%%%%%%%%%%%%%%%%%%%%%%%%%
\subsection{Future Research Directions}
\subsubsection{Dealing with Client Shift}
Domain shift between client datasets (client shift) has become a major concern of federated learning in medical image analysis.
To tackle this problem, domain adaptation~\citep{guan2022domain} has attracted extensive interest.
Classic domain adaptation methods typically need access to both source and target domains which may violate the privacy protection restraint in FL. 
Thus, developing more efficient federated domain adaptation methods will be a promising research direction. 
Another promising solution is personalized FL techniques~\citep{personalized1,personalized2} which utilize local data to further optimize a trained global model.

\subsubsection{Multi-Modality Fusion for FL}
Numerous imaging techniques/tools have been developed to create various visual representations of every subject, such as structural MRI, functional MRI, computed tomography (CT), and positron emission tomography (PET). 
Most existing FL studies only focus on images of a single modality.
How to leverage multi-modal imaging data in an FL system is an interesting problem with practical value.
Currently, a few works make early steps on FL with multi-modal medical data \citep{qayyum2022collaborative}.
More research work is expected on this topic in the future.

\subsubsection{Model Generalizability for Unseen Clients}
Most existing FL studies focus on model training and test within a fixed federation system.
That is, a global model is trained on and applied to the same client datasets (internal clients).
An interesting question is: When facing data from unseen sites that are outside of a federation (outside clients), how to guarantee the generalizability of an FL model? 
This is typically a domain generalization problem~\citep{DG-1,DG-2} or a test-time adaptation problem (\ie, using inference samples as a clue
of the unseen distribution to facilitate adaptation)~\citep{he2021autoencoder,varsavsky2020test}. 
Currently, there are a few works that introduce domain generalization into federated learning~\citep{jiang2023iop,liu2021feddg}.
In the future, evaluating and enhancing the generalizability of a trained FL model to unseen sites or even unseen classes (\ie, open-set recognition~\citep{geng2020recent,qin2022uncertainty}) will be a promising research direction.

\subsubsection{Weakly-Supervised Learning for FL}
Weakly-supervised learning is a promising technique that handles data with incomplete, inexact, and inaccurate labels.
These problems are common and widespread in medical imaging data.
How to deal with these ``imperfect" data (\eg, learning from noisy labels~\citep{karimi2020deep}) in an FL system is worthy of further exploration. 

%%%%%%%%%%%%%%%%%%%%%%%%%%%%%%%%%%%%%%%%%%%%%%%%%%%%%%%%%%%%%
\subsubsection{FL Security: Attack and Defense}
Several existing FL systems have been shown to be vulnerable to internal or external attacks, concerning system robustness and data privacy~\citep{lyu2022privacy}.
Further exploration of strong defense strategies in FL is helpful to enhance the security of FL systems. 
Another interesting question is: if an institution wants to withdraw from a federation, how to guarantee its data has been
removed from the trained FL model?
One solution is the data auditing technique~\citep{data-auditing} which can also be used to check if a poisoned/suspicious dataset is used in FL training.

%%%%%%%%%%%%%%%%%%%%%%%%%%%%%%%%%%%%%%%%%%%%%%%%%%%%%%%%%%%%%
\subsubsection{Blockchain and Decentralization of FL}
Most existing FL methods on medical tasks employ a centralized paradigm which demands a trustworthy central server.
This pattern gradually shows many disadvantages such as vulnerability to poisonous attacks and lack of credibility.
Recently, blockchain has been identified as a potentially promising solution to this problem~\citep{zhu2023blockchain}.
Using blockchain can avoid the dependence on the central server which can be the bottleneck of the whole federation.
Some work has made efforts on this point for medical image analysis through leveraging blockchain~\citep{kumar2021blockchain,noman2023blockchain} or other decentralization methods~\citep{roy2019braintorrent}. 
Currently, very limited work has been performed in this direction for medical image analysis, thus, there is much room for future research.

%%%%%%%%%%%%%%%%%%%%%%%%%%%%%%%%%%%%%%%%%%%%%%%%%%%%%%%%%%%%%%%%
\subsubsection{FL for Medical Video Analysis}
Most existing FL systems focus on combining cross-site medical images. 
As an extension of 2D/3D medical images, medical videos have been rarely explored.
Some pioneering work has employed FL to effectively take advantage of medical video from multiple sites/datasets for surgical phase recognition~\citep{kassem2022federated}.
In the future, FL systems consisting of medical videos for surgical or other applications will attract more research attention.

%%%%%%%%%%%%%%%%%%%%%%%%%%%%%%%%%%%%%%%%%%%%%%%%%%%%%%%%%%%%%%%%%%
\subsubsection{Large-Scale Medical Image Benchmark for FL}
Most existing medical image databases for FL research only consist of relatively small datasets for each client.
Some work just split a single large dataset (\eg, CheXpert~\citep{chexpert}) into different parts which are simulated as different client datasets. 
There is a lack of large-scale federations that include various sites across the world. 
Only a few works have leveraged real-world datasets from multiple cities or countries.
Li~\etal\citep{li2022integrated} collect chest X-ray images from different cities for COVID-19 detection.
Roth~\etal\citep{roth2020federated} leverage seven clinical institutions from across the world to build a federated learning model for breast density classification. 
Dayan~\etal\citep{dayan2021federated} build a large-scale federation through international cooperation.
Building large-scale benchmarks (including publicly available medical imaging databases and state-of-the-art FL algorithms) through extensive international cooperation is beneficial for FL applications in medicine.

%%%%%%%%%%%%%%%%%%%%%%%%%%%%%%%%%%%%%%%%%%%%%%%%%%%%%%%%%%%%%%%%%%
\subsubsection{Model Interpretability}
In clinical practice, one of the most significant hurdles in adopting machine learning and AI lies in the ``black box'' nature of certain machine learning systems, such as deep learning~\citep{vellido2020importance}. Even as FL emerges as a promising machine learning prototype, it encounters similar challenges. Thus, the issue of model interpretability remains a critical factor to address for the seamless integration of FL into clinical practice.
While some researchers have made an early effort towards this topic~\citep{li2023towards}, more explorations are expected in the future.

\subsubsection{Real-World Implementation and Practical Issues}
The majority of research on FL in medical imaging has primarily focused on algorithm development and simulation. 
FL methods, while promising, can encounter difficulties during real-world implementation, such as compatibility with existing hospital systems, integration challenges, and user adoption hurdles. Addressing these practical considerations is crucial for advancing the application of FL in medical image analysis.
%%%%%%%%%%%%%%%%%%%%%%%%%%%%%%%%%%%%%%%%%%%%%%%%%%%%%%%%%%%%%%%%%%%%%%%%%%%%%
\section{Conclusion} \label{Conclusion}
In this paper, we review the recent advances in federated learning (FL) for medical image analysis. We summarize existing FL methods from a system view and categorize them into client-end, server-end, and client-server communication methods. 
For each category, we provide a novel ``question-answer" paradigm to elaborate on the motivation and mechanism of different FL methods in medical image analysis.
We also introduce existing benchmark medical image datasets that have been used for federated learning.
In addition, we conduct an experiment to empirically compare representative FL methods on a popular benchmark imaging database (\ie, ADNI). 
We further discuss current challenges, potential research opportunities, and future directions of FL in medical image analysis.
We hope that this survey paper will provide researchers with a clear picture of the recent development of FL in medical image analysis and
that more research efforts can be inspired and initiated in this exciting research field. 

\section*{Acknowledgments}
This work was supported in part by NIH grants AG073297 and EB035160.
%This work was supported by NIH grant RF1AG073297. 
%%%%%%%%%%%%%%%%%%%%%%%%%%%%%%%%%%%%%%%%%%%%%%%%%%%%%%%%%%%%%%%%%%%%%%%%%%%%%%%%%%%%%%%%%%%%%%%%%%%%%
%\section*{Acknowledgment}
%H.~Guan, P.-T.~Yap, A.~Bozoki and M.~Liu were supported in part by NIH grant RF1AG073297. 
Part of the data used in this work were obtained from  Alzheimer's Disease Neuroimaging Initiative (ADNI). 
The investigators within ADNI contributed to the design and implementation of ADNI
and provided data but did not participate in the analysis or writing of this article. 

%%%%%%%%%%%%%%%%%%%%%%%%%%%%%%%%%%%%%%%%%%%%%%%%%%%%%%%%%%%%%%%%%%%%%%%%%%%%%%%
\section*{References}
\bibliography{refs}

\end{document}